\pdfoutput=1

\documentclass[11pt]{article}

\usepackage[final]{acl}

\usepackage{times}
\usepackage{latexsym}

\usepackage[T1]{fontenc}

\usepackage[utf8]{inputenc}

\usepackage{microtype}

\usepackage{inconsolata}

\usepackage{graphicx}

\usepackage{array}
\usepackage{fancybox}

\usepackage{fontawesome}

\usepackage{subcaption}  
\usepackage{caption}     

\usepackage{hyperref}

\usepackage{devanagari}     
\usepackage{CJKutf8}  
\usepackage{amsfonts}
\usepackage{amsmath}

\usepackage{enumitem} 

\usepackage[ ukrainian,greek, english]{babel}
\usepackage{cjhebrew}

\usepackage{algorithm}
\usepackage{algpseudocode}
\usepackage{xcolor}
\usepackage{framed}
\usepackage{mdframed}

\usepackage[symbol]{footmisc} 

\title{RomanLens: The Role of Latent Romanization in Multilinguality in LLMs}

\author{
 \textbf{Alan Saji\textsuperscript{1\footnotemark[2]\footnotemark[3]}},
 \textbf{Jaavid Aktar Husain\textsuperscript{2}\footnotemark[2]},
 \textbf{Thanmay Jayakumar\textsuperscript{1,3}},
 \textbf{Raj Dabre\textsuperscript{1,3,4,5}\footnotemark[4]}
 \\
 \textbf{Anoop Kunchukuttan\textsuperscript{1,6}},
 \textbf{Ratish Puduppully\textsuperscript{7}\footnotemark[3] } \\
 \\
 \textsuperscript{1}Nilekani Centre at AI4Bharat,
 \textsuperscript{2}Singapore University of Technology and Design,
 \\
 \textsuperscript{3}Indian Institute of Technology Madras, India,
 \\
 \textsuperscript{4}National Institute of Information and Communications Technology, Kyoto, Japan,
 \\
 \textsuperscript{5}Indian Institute of Technology Bombay, India,
 \textsuperscript{6}Microsoft, India,
 \textsuperscript{7}IT University of Copenhagen
\\
}

\usepackage{xcolor} 

\begin{document}

\maketitle
\begin{abstract}
Large Language Models (LLMs) exhibit strong multilingual performance despite being predominantly trained on English-centric corpora. This raises a fundamental question: \textit{How do LLMs achieve such multilingual capabilities?} Focusing on languages written in non-Roman scripts, we investigate the role of Romanization—the representation of non-Roman scripts using Roman characters—as a potential bridge in multilingual processing.
Using mechanistic interpretability techniques, we analyze next-token generation and find that intermediate layers frequently represent target words in Romanized form before transitioning to native script, a phenomenon we term \textit{Latent Romanization}. Further, through activation patching experiments, we demonstrate that LLMs encode semantic concepts similarly across native and Romanized scripts, suggesting a shared underlying representation. Additionally, for translation into non-Roman script languages, our findings reveal that when the target language is in Romanized form, its representations emerge earlier in the model’s layers compared to native script. 
These insights contribute to a deeper understanding of multilingual representation in LLMs and highlight the implicit role of Romanization in facilitating language transfer. Code and data are available at \href{https://github.com/AI4Bharat/Romanlens}{https://github.com/AI4Bharat/Romanlens}.
\end{abstract}
\section{Introduction}

The majority of modern Large Language Models (LLMs) \cite{touvron2023llama, dubey2024llama, team2024gemma} are trained predominantly on English-dominated corpora. Nonetheless, they exhibit strong multilingual performance across diverse languages \cite{shilanguage,huang-etal-2023-languages, zhao2024llama, zhang2023m3exam}. This raises a fundamental question: \textit{How do LLMs develop such robust multilingual capabilities despite their English-centric training?}

\begin{figure*}[t]
  \centering
  \includegraphics[width= \textwidth]{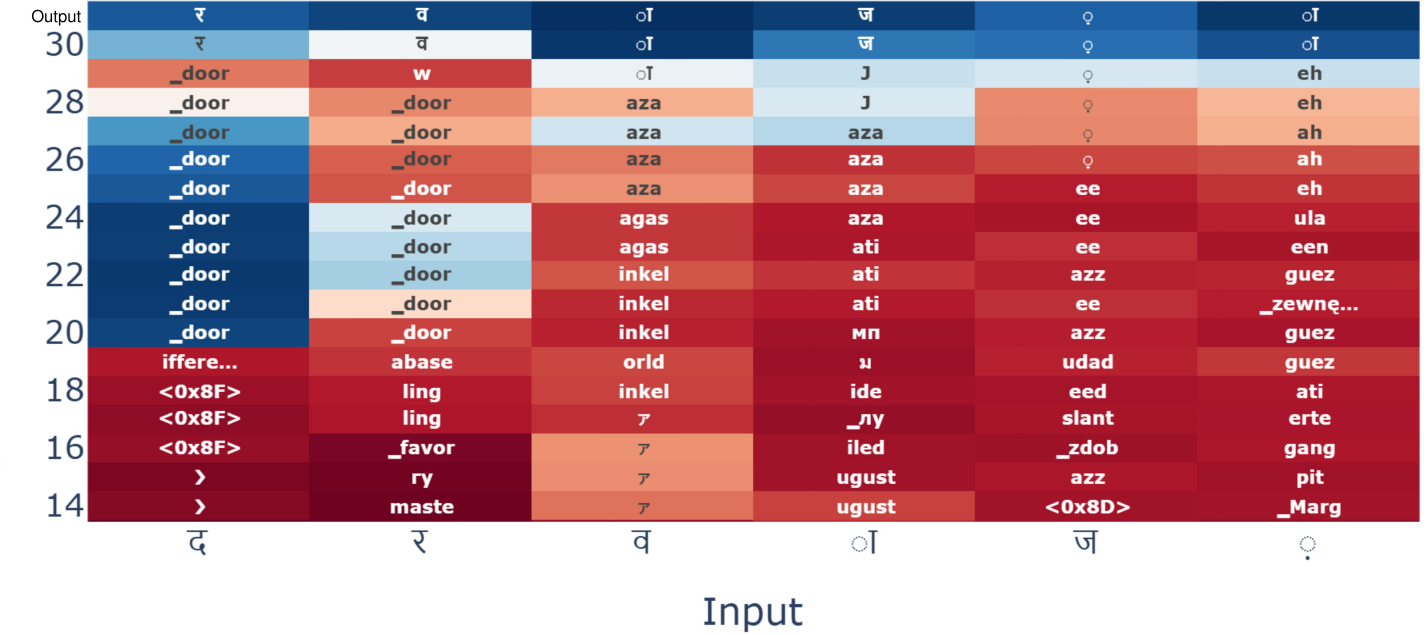}
    \caption{\textbf{Logit lens visualization} of  Llama-2 7B model 
    translating `door' from French to Hindi. We visualize the output ({\dn drvA)A}, `Darwaza' is the romanized form ) taking shape using logit lens producing a next-token distribution for each position (x-axis) and layers 14 and above (y-axis). Interestingly in the middle to top layers (20 - 29) we could observe romanized subwords of the Hindi word ({\dn vA} - w; {\dn a)A} - aza ;
{\dn )} -azz;   {\dn j} - j) and dependent vowels ( {\dn A} -a,eh ; 
 {\dn E} - i) before they are represented in their native script. Color represents entropy of next-token generation from low (blue) to high (red). Plotting tool: \cite{belrose2023eliciting}. 
}
    \label{fig:logit_lens_door}
\end{figure*}

\footnotetext[2]{Work done during employment at AI4Bharat.}
\footnotetext[4]{Work done during employment at NICT, Japan.}
\footnotetext[3]{\small{
\textbf{Correspondence:} Alan Saji (\href{mailto:alansaji2001@gmail.com}{alansaji2001@gmail.com})\\,  Ratish Puduppully
(\href{mailto:rapu@itu.dk}{rapu@itu.dk})
 }}

To address this, prior work by \citet{wendler-etal-2024-llamas} suggests that LLMs encode multilingual information within a shared, language-agnostic latent space, albeit with an inherent bias toward English due to training data composition and architectural choices. Building on this perspective, we investigate a complementary mechanism that may underlie multilingual processing, particularly for languages written in non-Roman scripts.


We hypothesize that LLMs leverage romanized forms of non-Roman script languages as an intermediate bridge between their language-agnostic concept space and language-specific output representations. Romanization—the representation of non-Roman scripts using Roman characters—may facilitate this process by aligning non-English languages more closely with English. Supporting this, \citet{husain2024romansetu} demonstrated that explicitly romanizing inputs improves model performance on multilingual tasks, suggesting an inherent alignment between romanized text and English representations. We investigate whether LLMs indeed use romanization as a bridge between language-agnostic concepts and language-specific outputs given its potential implications for understanding multilingual processing in LLMs.

Our primary experiment visualizes next-token generation using the logit lens \cite{logit-lens}, applying the language modeling head to intermediate layers. As illustrated in Figure \ref{fig:logit_lens_door}, we prompt the LLaMA-2 7B \cite{touvron2023llama} model with ``Francais: porte - {\dn Eh\306wdF}:'' to translate ``door'' from French to Hindi. Our results show that in the middle-to-top layers (layers 20–29), romanized Hindi subwords intermittently appear before transitioning to native script, suggesting an internal representation of romanized text as an intermediary. Additionally, these romanized representations increase in prominence across timesteps as the target word is generated.

To further probe this phenomenon, we employ activation patching \cite{ghandeharioun2024patchscope, variengien2024look, chen2024selfie, dumas2024separating}, a technique that replaces activations from one forward pass with another to analyze the resulting outputs (c.f. Section \ref{subsection: activation patching explained}). \citet{dumas2024separating} found that LLMs process language and conceptual information as distinct entities. Building on this, we perform layerwise activation patching between romanized and native-script inputs to examine whether LLMs encode conceptual information similarly across scripts.

\vspace{0.5em}

\begin{figure}[t]
\centering
\includegraphics[width= 0.8\columnwidth]{latex/fig/pushpam_fig3_3.pdf}

\caption{\textbf{Translation comparison: Romanized vs. Native script.} Next-token generation is visualized using the logit lens for the Gemma-2 9B model translating "flower" from French to Malayalam in romanized (left) and native script (right). The x-axis shows next-token distributions; the y-axis covers layers 30 and above. Target language representations (e.g.,``push",``poo") appear 1–2 layers earlier in romanized outputs compared to native script outputs.( \raisebox{-0.5ex}{\includegraphics[height=2ex]{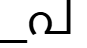}}, 'pa' is its romanized representation). \textit{Push} is a prefix of \textit{pushpam}, the romanized form of the translation of ``flower''; also, \textit{poo} is another romanized translation of ``flower''.Color represents entropy of next-token generation from low (blue) to high (red). Plotting tool: \cite{belrose2023eliciting}.
    }
    \label{fig:flower romanized vs native}
\end{figure}

\noindent Based on our experiments, we list down a summary of our contributions below:

\noindent\textbf{1. Latent Romanization Across Layers:} During multilingual next-token generation, intermediate layers occasionally represent tokens in Romanized form before resolving to native script (Figure \ref{fig:logit_lens_door}). We term this phenomenon \textit{Latent Romanization}.

\noindent\textbf{2. Consistent Semantic Encoding Across Scripts:} Activation patching experiments reveal that LLMs encode semantic concepts similarly, regardless of whether the input is in native or Romanized script.

\noindent\textbf{3. Earlier Emergence of Target Representations:} When translating into Romanized versus native script, Romanized target representations emerge earlier in the model’s layers—typically one or two layers prior to native script representations (c.f. Figure \ref{fig:flower romanized vs native}). 


   \section{Related Work}

Recent studies have explored various aspects of LLMs' multilingual behavior: examining whether English emerges as a latent language in English-centric LLMs \cite{wendler-etal-2024-llamas}, how the composition of training corpus mixtures influences latent representations \cite{zhong2024beyond} and how LLMs handle multilingual capabilities \cite{zhao2024large}. \citet{kojima-etal-2024-multilingual} describe distinct phases in multilingual information processing: initial layers map language-specific lexical and syntactic representations to a language-independent semantic space, middle layers maintain this semantic abstraction, and final layers transform these representations into language-specific lexical and syntactic forms. Interpretability tools relevant to this work include logit lens \cite{logit-lens}, tuned lens \cite{belrose2023eliciting} and direct logit attribution \cite{elhage2021mathematical} which are key tools for decoding intermediate token representations in transformer models. The logit lens applies the language modeling head to earlier layers without additional training, while the tuned lens in addition to this trains an affine mapping to align intermediate states with final token predictions. Direct logit attribution attributes logits to individual attention heads. This work focuses on the logit lens (Section \ref{sec:latent_romanization_analysis}) to investigate whether English-centric decoder only LLMs when prompted in a non-Roman script language, processes via romanized latent states before producing native language text. Tuned lens is avoided as its training process might obscure the intermediate romanized states by aligning them to final native script outputs, potentially masking the phenomenon under investigation.

Activation patching \citep{meng2022locating} is a key interpretability technique employed in our study. This technique has been used to draw causal interpretations of LLMs representations \cite{variengien2024look,geiger2022inducing,kramar2024atp,ghandeharioun2024patchscope,chen2024selfie}. Building on these approaches, we adopt an activation patching-based experimental framework to investigate and compare how concepts are encoded in romanized versus native scripts.

Previous studies have demonstrated that romanization can serve as an effective approach to interact with LLMs \cite{husain2024romansetu}. \citet{liu-etal-2024-translico} and \citet{xhelili-etal-2024-breaking} employ an approach based on contrastive learning for post-training alignment, contrasting sentences with their transliterations in Roman script to overcome the script barrier and enhance cross-lingual transfer.

However, our work distinguishes itself from prior research by exploring the presence of romanized representations in the latent layers of an LLM during multilingual tasks, an aspect that, to the best of our knowledge, has not yet been investigated.

\section{Background}
We give a quick background of the transformer's forward pass, romanization and the basics of mechanistic interpretability approaches such as logit lens and activation patching which we leverage in this paper.

\paragraph{Transformer's Forward Pass}
Decoder-only transformer models (Vaswani, 2017) employ a residual architecture to process input sequences through multiple layers, producing a sequence of hidden states (latents). These latents, whose dimensionality remains the same, are updated iteratively across layers through transformer blocks $f_j$, where $j \in [0, k]$ indicates the layer index and $k$ is the final layer index. For next-token prediction, the final latent $h^{(k)}_i$ is transformed by an unembedding matrix $U \in \mathbb{R}^{v×d}$ to produce logit scores for vocabulary tokens which are then converted to probabilities via the softmax function (c.f. Appendix \ref{sec:detailed_transformers_forward_pass}).

\subsection{Romanization}
\label{sec:romanization}
Transliteration is the conversion of text written in one script to another. Romanization is a subcategory of transliteration where the target script is English/Latin. Within romanization there are multiple romanization schemes available, each based on different considerations. One key aspect of romanization schema is if it is lossy or lossless. A lossless scheme is required in cases where we have to convert the output back to native script. Typically, deterministic transliterations are lossless, whereas natural transliterations are lossy. 

\paragraph{Example:}
The Hindi word for “flower” in Devanagari and its romanization:

\begin{itemize}
\item \textit{Devanagari} (native script): {\dn \8{P}l}
\item \textit{Romanization}: phool
\end{itemize}

\subsection{Interpretability Tool: Logit lens}
Generally, in a decoder only LLM, the unembedding matrix $U$ is multiplied with the final hidden state and a softmax is taken on the product to produce the token distributions at that token generation step. Since all hidden states of an LLM are in the same shape, 
it is possible to apply the unembedding matrix and softmax on all layers, thereby generating token distributions at all layers.
This method of prematurely decoding hidden states is referred to as 
\textit{logit lens} \cite{logit-lens}. Logit lens reveals how the latent representations evolve across layers to produce the final output, providing insights into the progression of computations within the model.

\subsection{Interpretability Tool: Activation Patching}
\label{subsection: activation patching explained}
Activation patching involves modifying or patching the activations at specific layers during a forward pass and observing the effects on the model's output. In this work, we adopt the activation patching setup introduced in \citet{dumas2024separating}.

In the context of activation patching, let $\ell$ denote the language of a word, $C$ denote the concept of a word and $w(C^\ell)$ denote that word. For example, if $C$ = cow and $\ell$ = `en', then $w(C^{en})$ = `cow'. Similarly $w(C^{fr})$ = `vache'. We use 5-shot translation prompts to create paired source  \( S = ( C_S, \ell_{S}^\text{in}, \ell_{S}^\text{out}) \) and target prompt \( T = (C_T, \ell_{T}^\text{in}, \ell_{T}^\text{out}) \), with different concept, input language, and output language. Unless otherwise specified, \( \ell_S \) and \( \ell_T \) refer to the output languages of \( S \) and \( T \), respectively.

\begin{figure}[t]
    \centering
    
    \includegraphics[width=\columnwidth]{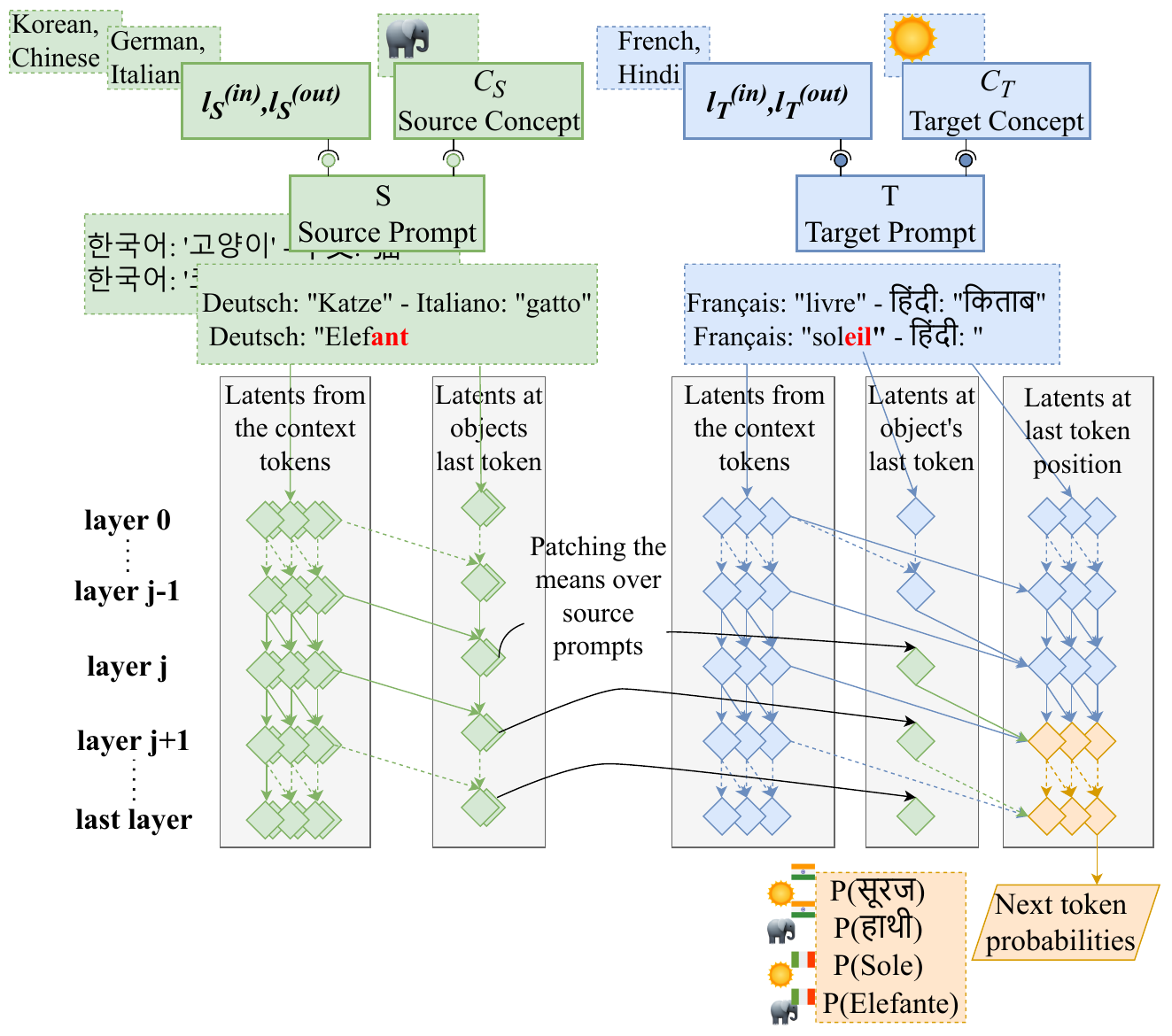} 
    \caption{ 
    \textbf{Activation patching illustration.}
    For two given concepts, say, \textit{elephant} and
\textit{sun}, we generate multiple source prompts which
translate \textit{elephant}, and a target prompt for translating \textit{sun}
from French to Hindi. We then extract the residual stream associated with the final token of the word to be translated after a specific layer \textit{j} and all subsequent layers from the source prompts. The mean residuals at each layer are computed and inserted into the corresponding positions during the forward pass of the target prompt. The resulting next token probabilities will
be dominated by the source concept in target language
(ELEPHANT\textsuperscript{HI}, i.e.,
{\dn hATF}) when patching at layers 0–15, and by
the target concept in target language (SUN\textsuperscript{HI}, i.e.,
 {\dn \8{s}rj}) for layers 16–31. Adapted from \citet{dumas2024separating}.}

 \label{fig: activation_patching_illustration} 
    
\end{figure}

For each transformer block $f_j$, we create two parallel forward passes: one processing the source prompt 
$S = (s_1, \dots, s_{n_s},\dots,s_{n_S})$ and the other processing the target prompt 
$T = (t_1, \dots,t_{n_t},\dots, t_{n_T})$. It should be noted that ${n_s}$, ${n_t}$ represents the position of the last token of the object to be translated whereas
 ${n_S}$, ${n_T}$ represent the last token position of the source and target prompt to be translated. In Figure \ref{fig: activation_patching_illustration} in the target prompt translating \textit{sun} from French to Hindi, ${n_t}$ would be the position of the subword $``eil"$ highlighted in red, whereas ${n_T}$ would be the position of ``{\dn dF}" the last subword of the prompt. Similarly, for the source prompt in Figure \ref{fig: activation_patching_illustration} translating \textit{elephant} from German to Italian both ${n_s}$ and ${n_S}$ would be the position of  $``ant"$ highlighted in red. After creating the parallel forward passes, we extract the residual
stream of the last token of the word to be translated at ${n_s}$
after layer $j$, denoted as $h^{(j)}_{n_s}(S)$ and all subsequent layers, and insert it at the corresponding layer at the corresponding position 
$n_t$ in the forward pass of the target prompt, i.e. by setting 
$h^{(j)}_{n_t}(T) = h^{(j)}_{n_s}(S), 
h^{(j+1)}_{n_t}(T) = h^{(j+1)}_{n_s}(S),\dots, h^{(k)}_{n_t}(T) = h^{(k)}_{n_s}(S) $.
We then complete the altered forward pass and analyze the next token distribution to evaluate source concept $C_S$  encoded in the target language. An illustration of this setup is shown in Figure \ref{fig: activation_patching_illustration}.

\section{Methodology}
\label{sec:methodology}



We design our analysis setup with the intention of addressing the following research questions:

\noindent\textbf{RQ1:} Do LLMs exhibit latent romanization during multilingual text completion tasks? (Section \ref{sec:latent_romanization_analysis})

\noindent\textbf{RQ2:} How does the representation of semantic concepts in LLMs compare between native and romanized scripts of non-Roman script languages? (Section \ref{sec:activation_patching})

\noindent\textbf{RQ3:} What are the differences in hidden layer representations when processing the same language in romanized and native scripts? (Section \ref{sec:translate_roman_native})
    



\vspace{0.5em}

\noindent\textbf{Prompt design.} We design prompts that facilitate next-token ($x_{n+1}$) prediction from the given context ($x_1, \ldots, x_n$). 
This is adopted across all analysis setups. The prompts are designed around translation, repetition, and cloze tasks, as described below.

\vspace{0.25em}
\noindent\textit{Translation task.} We prompt the model to translate a word given five in-context examples.



 
\vspace{5pt} 
\noindent\textit{Repetition task.} We prompt the model to repeat a word in the same language given five in-context examples. 

\vspace{5pt}

\noindent\textit{Cloze task.} We prompt the model to predict the masked word in a sentence given two in-context examples.

\vspace{0.25em}

These tasks cover a range of multilingual text completion setups. Among these, the repetition task is more syntactic in nature compared to translation and cloze tasks. Appendix \ref{sec:English_Translation_and_Romanization_of_the_Sample_Prompts} provides a Hindi example of prompts across each of the three tasks, along with their English translations and romanized forms.


    


 \subsection{Latent Romanization Analysis}
 \label{sec:latent_romanization_analysis}
 Translation, repetition, and cloze tasks are explored by providing the respective prompts as inputs to an LLM to generate the corresponding output word. We romanize the output word, tokenize it, retaining only the tokens present in the model's vocabulary, and analyze the occurrence of these tokens in the latent layers across timesteps of the output word generation. The analysis is done using logit lens by examining whether the probability of a romanized token in the next token distribution at a given layer exceeds 0.1. We refer to this hereafter as the \textit{latent romanization condition}. The 0.1 threshold is empirically determined to optimize detection accuracy, i.e. minimizing false positives and maximizing true positives (compared to alternative thresholds 0.05 and 0.01). Our analysis focuses on the final 10 layers of an LLM, where coherent romanized representations emerge according to logit lens visualizations (c.f. Figures \ref{fig:logit_lens_door} and \ref{fig:logit_lens_love_greek} - \ref{fig:logit_lens_amharic_music}).


 We track romanized tokens using a timestep-specific tokenization scheme optimized for detection accuracy. In the first output generation timestep, we check for tokens that include the full romanized word and its prefixes. During intermediate timesteps, we check for all possible substrings of the romanized word in the latent layers. In the final output generation timestep, we probe the presence of only the full romanized word and its suffixes as potential tokens (c.f. Appendix \ref{sec:latent_romanization_tokenization_scheme}).
 
 Latent romanization is analyzed under three distinct scenarios:

 \begin{enumerate}[label=(\alph*)]

 \item \textbf{Constrained Word Generation}: Using standard prompts and the target word, we guide the model to generate the complete target word. At each layer, we track how often romanized tokens emerge during the decoding process. We do this by checking each generation timestep for \textit{latent romanization condition}. The \textit{`latent fraction'} for a layer represents how frequently these romanized tokens  appear across timesteps, averaged across all samples (c.f. Appendix \ref{sec:latent_fraction}). 
 \item \textbf{First Subword Only}: We prompt for only the initial subword and compute the \textit{latent fraction}. Despite having a single timestep, we maintain the \textit{latent fraction} terminology for consistency.

 \item \textbf{Last Subword Only}: We augment the standard prompt with all but the final subword of the target, then analyze the generation of the final subword.

\end{enumerate}

We document layerwise \textit{latent fraction} separately for first and last subword generation of the output. Intuitively, there is a distinction between the first and last token generation steps for a given word. In the former, the model faces a greater decision-making burden, while in the latter, the model is typically more confident in its predictions. We hypothesize that, in the latter scenario, the model may reach a decision in the layers just below the final few layers and express the output in a romanized form, as language-specific neurons, responsible for native script processing, are concentrated primarily in the last few layers \cite{tang-etal-2024-language}. This could lead to romanized tokens appearing more frequently as the model progresses from the first subword to the last.

 \subsection{Patching With Romanized Representation Versus Native Representation}
 \label{sec:activation_patching}
 
 We intend to compare how concepts are encoded in native script versus romanized script using translation task. In order to do this, we patch representations from source prompt where input language is romanized and compare this with patching from source prompt where input language is in native script. We first perform patching using a single-source prompt. Then, we repeat the process using an averaged multi-source prompt, contrasting multiple romanized source input languages with multiple native script source input languages. Single-source patching might be influenced by language or script-specific characteristics. In contrast, multi-source patching reduces such biases, leading to more robust and generalizable findings. 
  
In all scenarios from the resulting next token distribution, we compute the probabilities 
$P(C^{\ell_T}_S)$ i.e. probability of source concept in target language, and $P(C^{\ell_T}_T)$ i.e. probability of target concept in target language. We track $P(C^{\ell})$, i.e., the probability of
the concept C occurring in language $\ell$, by simply
summing up the probabilities of all prefixes
of $w(C^{\ell})$ and its synonyms in the next-token distribution (c.f. Appendix \ref{sec:Computing_Probability_Activation_Patching_Experiment}). We analyze  $P(C^{\ell_T}_S)$  to evaluate how effectively a concept is encoded in a given source input language $\ell_{S}^\text{in}$.

\begin{figure*}[t]
    \centering
        \hfill
    \begin{subfigure}{0.32\textwidth} 
        \centering
        \includegraphics[width= \textwidth]{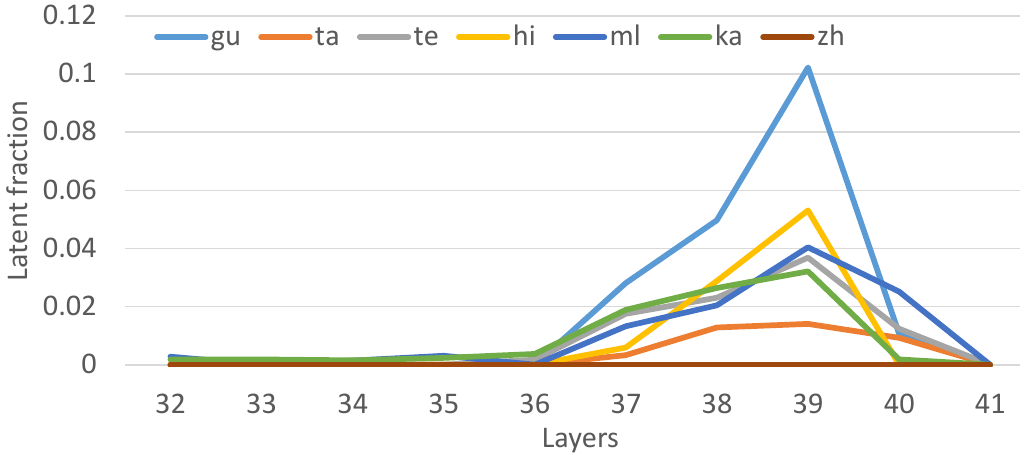} 
        \caption{All generation steps.}
        \label{fig:all tokens latent fraction}
    
    \end{subfigure}
     \hfill
    \begin{subfigure}{0.32\textwidth} 
        \centering
        \includegraphics[width= \textwidth]{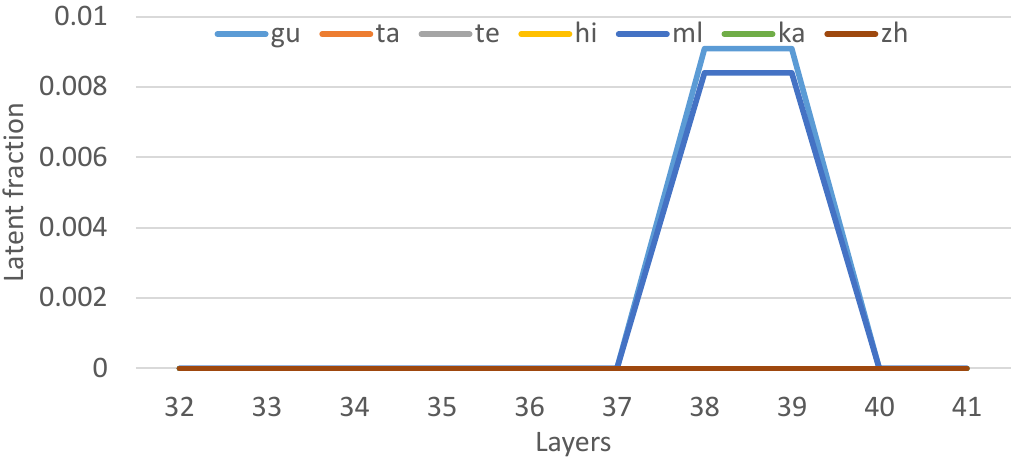} 
        \caption{First token generation step. }
        \label{fig:first_token_layerwise_lat_rom}
    \end{subfigure}
    \hfill
    \begin{subfigure}{0.32\textwidth} 
        \centering
        \includegraphics[width= \textwidth]{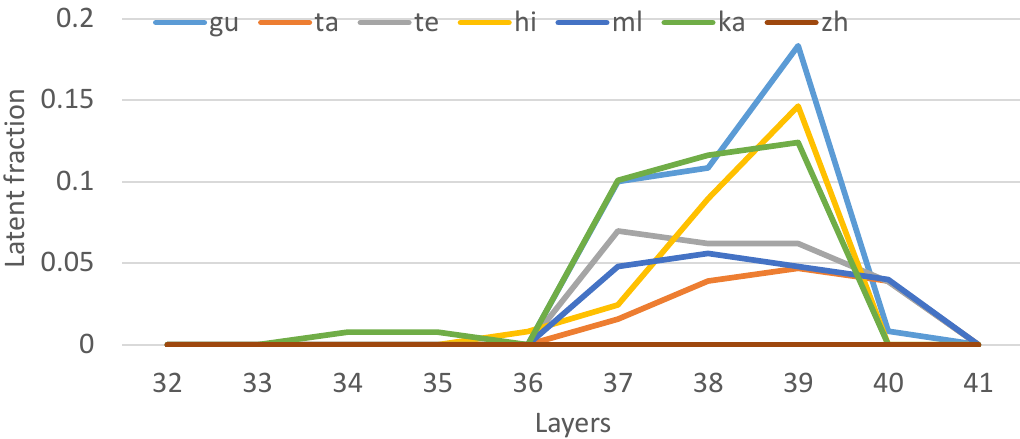} 
        \caption{Last token generation step.}
        \label{fig:last_token_layerwise_lat_rom}
    
    \end{subfigure}
     \hfill
     \caption{\textbf{Distribution of Romanized Tokens Across Model Layers: Analysis of First, Last, and All Generation Timesteps.} This distribution is plotted across the last 10 layers of Gemma-2 9b IT model for translation task with English as source language and is averaged across 100+ samples. \textit{X}-axes represents layer index, \textit{y}-axes represents latent fraction i.e. the fraction of timesteps where romanized tokens occur with a probability > 0.1 averaged over samples for a specific layer.  We plot the distributions for Gujarati (gu), Tamil (ta), Telugu (te), Hindi (hi), Malayalam (ml), Georgian (ka) and Chinese (zh).}
                
\label{fig:latent_romanization first and last token}
\end{figure*}

 \subsection{Comparing Translations Into Romanized vs. Native Script}
 \label{sec:translate_roman_native}
This analysis examines translation task with target languages in native script and their romanized equivalents. We focus on first-token generation of the output word, also considering possible synonyms.

 In the next token generation step, the probability of target language and latent language (English) \cite{wendler-etal-2024-llamas} at each layer is examined using logit lens. Each probability is computed by summing over probabilities of all possible tokens corresponding to the answer word(s) in that respective language (c.f. Appendix \ref{sec:Computing_Language_probabilities}). Tokens of latent language and target language are derived using tokenization scheme for first token generation timestep mentioned in Section \ref{sec:latent_romanization_analysis}.

\section{Experimental Settings}

\paragraph{Languages:}
\label{sec: languages}
We focus on five Indic languages: Hindi, Gujarati, Tamil, Telugu, and Malayalam, as well as Chinese and Georgian. Among these, Hindi and Gujarati belong to the Indo-Aryan branch of the Indo-European language family and use scripts derived from the Devanagari and Gujarati scripts, respectively. Tamil, Telugu, and Malayalam, on the other hand, are part of the Dravidian language family and each has its own distinct script. Chinese belongs to the Sino-Tibetan language family and is written using logographic characters. Georgian is part of the Kartvelian language family and uses the unique Georgian script. To examine the generality of latent romanization, we perform qualitative analyses on five additional languages that use different writing systems: Greek, Ukrainian, Amharic, Hebrew, and Arabic
(c.f Appendix \ref{sec:logit_lens_qualitative_analysis_appendix}).

\begin{figure}[t]
    \centering
    \includegraphics[width=0.45\textwidth]{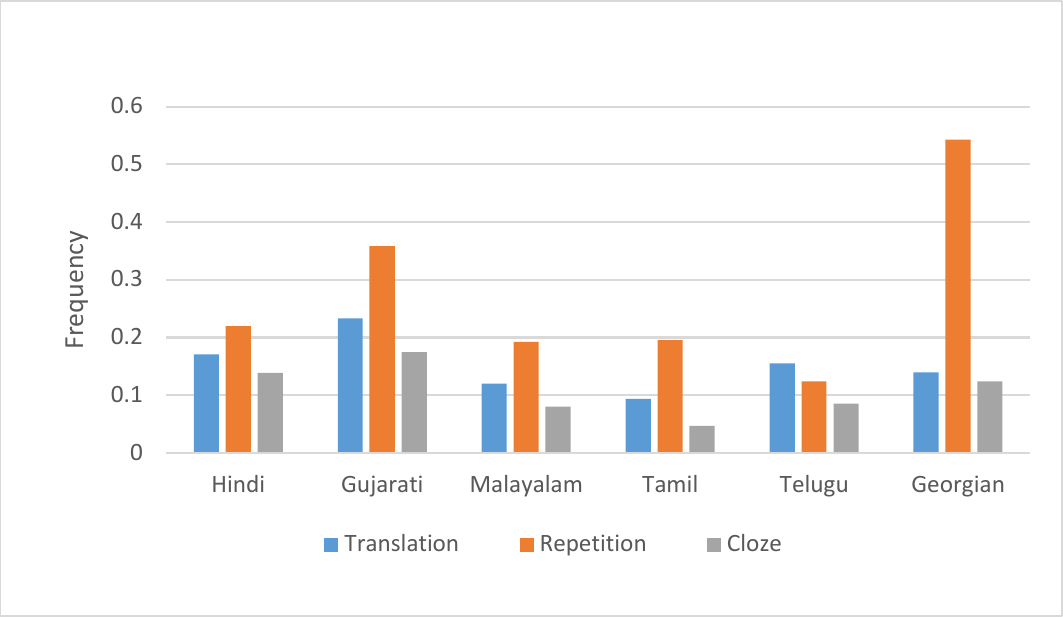}
    \caption{\textbf{Frequency distribution of romanized tokens across translation, repetition and cloze task.} We check if romanized tokens occur with a probability > 0.1 in the last 10 layers of an LLM and compute frequency of this occurrence across 100+ samples. Gemma 2 9B IT is the model used and English is the source language for translation task.}
    \label{fig:last token romanization freq}
\end{figure}

\begin{figure*}[t]
    \centering
    \begin{tabular}{@{}p{6.5cm}p{6.5cm}p{1cm}@{}}
        \multicolumn{1}{c}{\textbf{Native}} & 
        \multicolumn{1}{c}{\textbf{Romanized}} & 
        \multicolumn{1}{c}{\textbf{$D_{KL}$}} \\ 
        [1ex]
        
        \begin{minipage}{0.36\textwidth}
            \centering
            \includegraphics[width = \columnwidth]{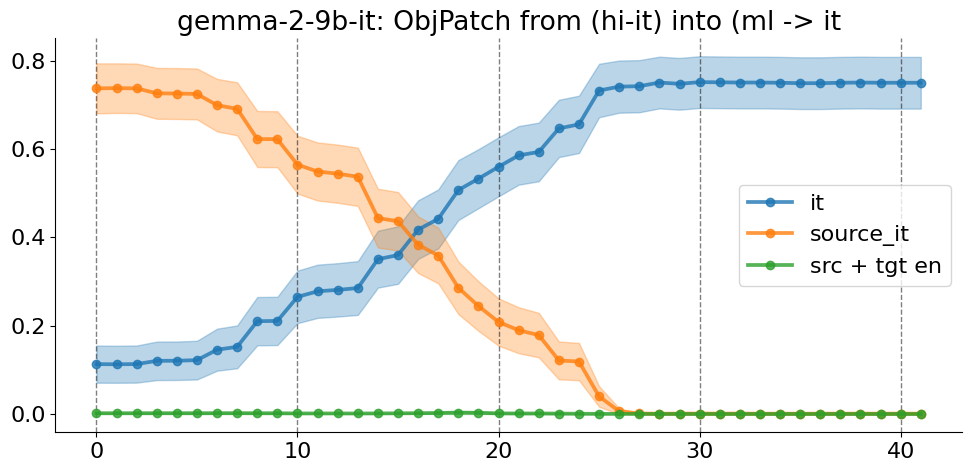}
        \end{minipage} &
        \begin{minipage}{0.36\textwidth}
            \centering
             \includegraphics[width = \columnwidth]{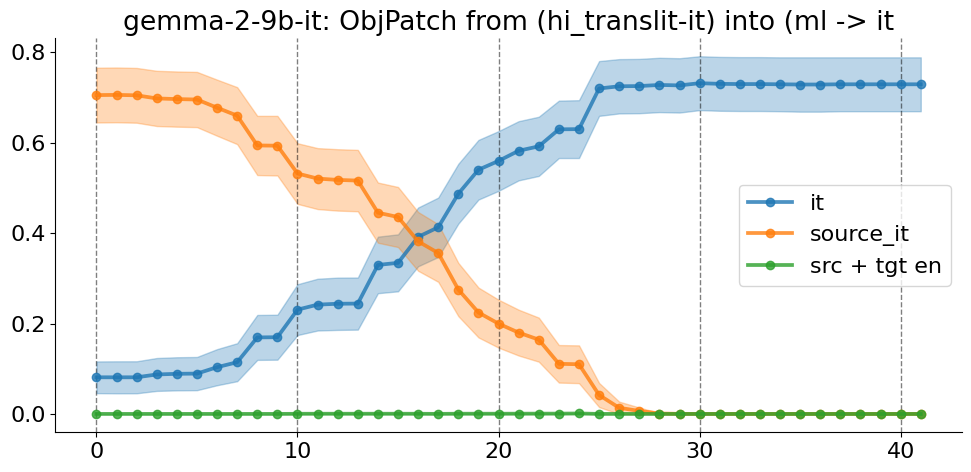}
        \end{minipage} &
        $0.0006$ \\
        \multicolumn{2}{c}{\textbf{Single Source}} & \\
        [2ex]
        
        \begin{minipage}{0.36\textwidth}
            \centering
            \includegraphics[width = \columnwidth]{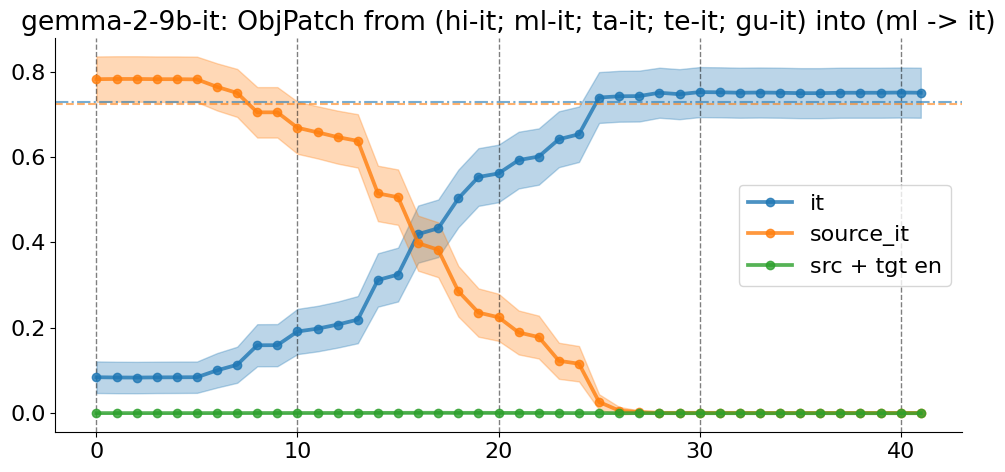}
        \end{minipage} & 
        \begin{minipage}{0.36\textwidth}
            \centering
            \includegraphics[width = \columnwidth]{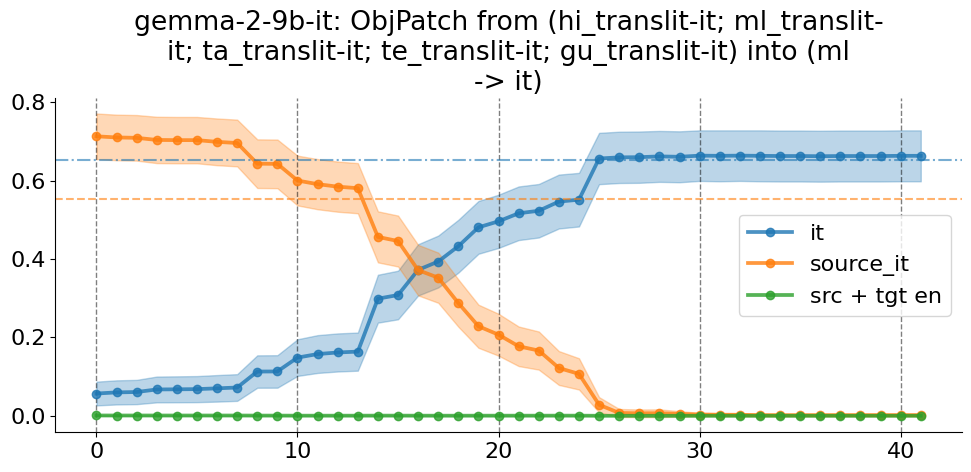}
        \end{minipage} & 
        $0.001$ \\
        \multicolumn{2}{c}{\textbf{Multi Source}} & \\
    \end{tabular}
    \caption{\textbf{Comparative Analysis of Patching from Source Prompts: Native Script vs. Romanized Script Inputs.} Concept probabilities across layers for different prompt setups are plotted in each graph. The x-axis represents the patching layer, while the y-axis indicates the probability of correctly predicting the concept in language $\ell$. Curves: blue (target concept in Italian), orange (source concept in Italian), and green (source or target concept in English). Results are reported as means with 95\% Gaussian confidence intervals, calculated over a dataset of 200 samples. The orange curve is compared across adjacent graphs and KL divergence $D_{KL}$ quantifies this. Languages involved: Hindi (hi), Tamil (ta), Telugu (te), Malayalam (ml), Gujarati (gu) and Italian (it). Model: Gemma 2 9b it.}
    \label{fig:patching from mean and source prompts}
\end{figure*}

\begin{figure}[t]
    \centering
    \begin{subfigure}{0.45\columnwidth}
        \centering
        \includegraphics[width=\textwidth]{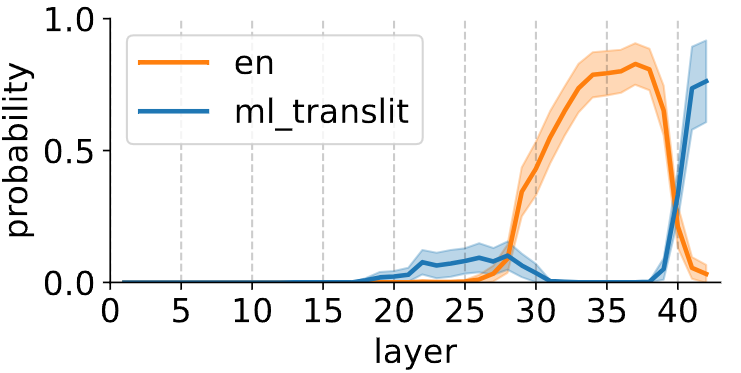}
        \caption{fr → ml (romanized)}
        \label{fig:fr-ml_translit}
    \end{subfigure}
    \hfill
    \begin{subfigure}{0.45\columnwidth}
        \centering
        \includegraphics[width=\textwidth]{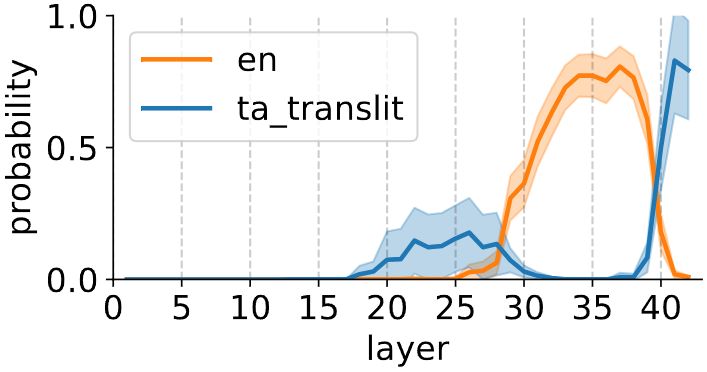}
        \caption{fr → ta (romanized)}
        \label{fig:fr-ta_translit}
    \end{subfigure}
    
    \begin{subfigure}{0.45\columnwidth}
        \centering
        \includegraphics[width=\textwidth]{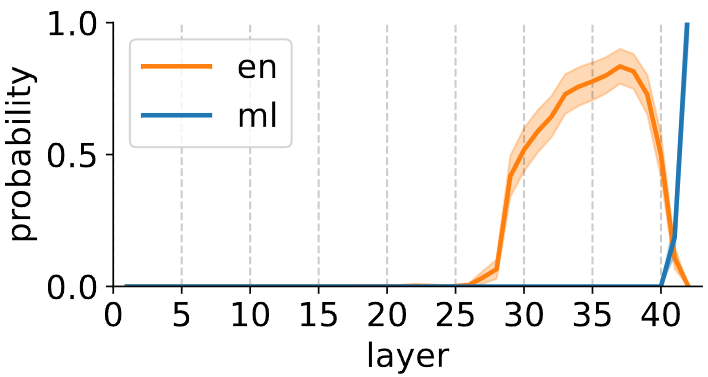}
        \caption{fr → ml (native)}
        \label{fig:fr-ml}
    \end{subfigure}
    \hfill
    \begin{subfigure}{0.45\columnwidth}
        \centering
        \includegraphics[width=\textwidth]{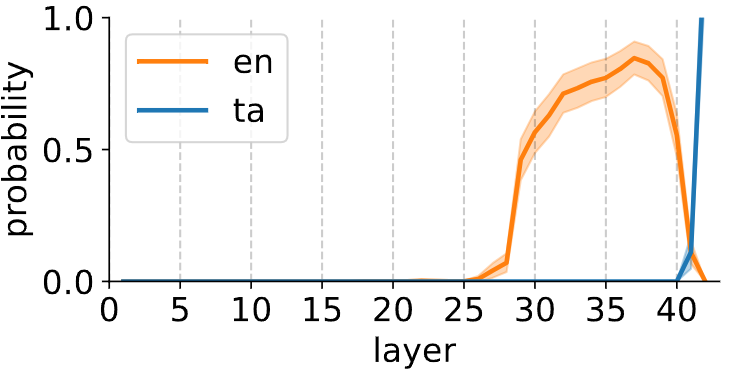}
        \caption{fr → ta (native)}
        \label{fig:fr-ta}
    \end{subfigure}

    \caption{\textbf{Language probabilities for latent layers} in translation from French to Malayalam and Tamil  in romanized (top row) and native scripts (bottom row) across various samples  using Gemma-2 9B IT model. X-axis: layer index; Y-axis: probability of correct next token (via logit lens) in a given language. Error bars: 95\% Gaussian confidence intervals. English is the latent language (orange curve). For romanized script, target representations (blue curve) emerge 1-2 layers earlier than native script, appearing before layer 40.}
    \label{fig:romanization early}
\end{figure}

\paragraph{Language Models:}
In this study, we focus mainly on Gemma-2 9B, Gemma-2 9B instruction-tuned \cite{team2024gemma}, Llama-2 7B, Llama-2 13B \cite{touvron2023llama} and Mistral-7B \cite{jiang2023mistral} (c.f. Appendix \ref{sec:mistral}) language models, some of the best performing open weights English-centric LLMs. Although the training data for these models are primarily English, these models have high multilingual capabilities \cite{huang-etal-2023-languages,zhao2024llama,zhang2023m3exam}.


\paragraph{Romanization:}
 We have used the IndicXlit scheme \cite{madhani-etal-2023-aksharantar} (c.f. Appendix \ref{sec:romanization_scheme_indic_languages}) for Indic languages, \textit{pypinyin} \cite{pypinyin} for Chinese and \textit {Unidecode} \cite{unidecode} for Georgian to romanize native scripts.

\paragraph{Data For Logit Lens Experiments:}
\label{sec:logit_lens_data}
 We use a curated word-level dataset with synonyms translated from recent works in this field \cite{wendler-etal-2024-llamas} using the Llama 3.3 70B model \cite{dubey2024llama}. The quality of translations were manually verified to ensure accuracy and relevance. The datasets are kept simple to facilitate the observation of how latents evolve at each token level.

\paragraph{Data For Activation Patching Experiments:}
\label{sec:activation_patching_data}
We adopt the dataset used in recent studies \cite{dumas2024separating}, extending it to include both native script and romanized versions of the languages considered in this
study. Translations are performed using the Llama 3.3 70B model \cite{dubey2024llama} and the translations are romanized using IndicXlit \cite{madhani-etal-2023-aksharantar}, \textit{pypinyin} \cite{pypinyin} and \textit {Unidecode} \cite{unidecode}. All translations were manually validated to ensure data quality.

\section{Results}

\subsection{Latent romanization}
Our analysis demonstrates that \textbf{LLMs do exhibit latent romanization during text completion tasks in six out of seven quantitatively analyzed languages} (c.f. Figure \ref{fig:all tokens latent fraction}), with Chinese being the only exception where this phenomenon is not observed. In Figure \ref{fig:all tokens latent fraction} we can see the \textit{latent fraction} of romanized tokens for the last 10 layers of an LLM. This is across all tokens of the output word. The frequency of romanized tokens tends to increase just before the last layers. Qualitative logit lens analysis done for Greek, Ukrainian, Arabic, Amharic and Hebrew reveals similar patterns (c.f. Appendix \ref{sec:logit_lens_qualitative_analysis_appendix}). 

In Figures \ref{fig:first_token_layerwise_lat_rom} and \ref{fig:last_token_layerwise_lat_rom} it is observed that the range of the \textit{latent fraction} of romanized tokens varies from $0 - 0.01$ in the first token generation step to $0 - 0.2$ in the last token generation step in most languages. This trend indicates that latent romanization increases progressively from the initial token to the final token of the output across languages. This observation supports our hypothesis that the first token generation involves more intricate decision-making processes compared to the generation of the final token within an output word.


  Based on the above information, we quantify romanization across tasks for the last token generation step. Here the criteria would be if a romanized token occur at next token generation step with a probability > 0.1 in the last 10 layers, it is counted as a positive. As depicted in Figure \ref{fig:last token romanization freq}, across translation, repetition and cloze task we observe a significant occurrence of romanized tokens in the latent layers. 

  Among the three text completion tasks considered in the latent romanization experiment (Section \ref{sec:latent_romanization_analysis}), a relatively high latent romanization frequency is observed for the repetition task (Figure \ref{fig:last token romanization freq}). The translation and cloze tasks function at a semantic level, whereas the repetition task is purely syntactic. This means that the repetition task being less complex, the model may reach a decision of what to predict sooner, potentially in earlier layers and might express its prediction in romanized form in intermediate layers. This behavior could be attributed to language-specific neurons, responsible for native script processing, being predominantly concentrated in the initial and final few layers of LLMs, \cite{tang-etal-2024-language} leaving the intermediate layers without them.

\subsection{Patching With Romanized Representation vs. Native Representation}
In Figure \ref{fig:patching from mean and source prompts}, we analyze two patching scenarios: In the single-source setup, we compare patching from Hindi→Italian with Hindi(romanized)→Italian source prompt, to Malayalam→Italian target prompt. In the multi-source setup, we contrast patching from multiple native script prompts (Hindi→Italian, Gujarati→Italian $\ldots$) against their romanized counterparts (Hindi(romanized)→Italian, Gujarati(romanized)→Italian $\ldots$). We compare the probability distributions of source concept in target language $P(C^{\ell_T}_S)$ across adjacent graphs where native source input language is contrasted with romanized source input language. It is evident that these probability distributions show remarkable similarity whether the source input language is in romanized or native script, consistent across both single-source and multi-source prompt setups. The similarity is quantitatively supported by the KL divergence measurements between adjacent graphs, remaining below 0.01 in both setups. KL divergence value close to zero indicates that the two distributions are nearly identical. 

This analysis reveals that \textbf{LLMs encode semantic concepts similarly regardless of whether the input is in native or romanized script}. Furthermore, this finding demonstrates that the model achieves comparable levels of language understanding when processing non-Roman script languages in their romanized form as in their native script. 


\subsection{Comparing Translations Into Romanized vs. Native Script}

 We quantify the observations from Figure \ref{fig:flower romanized vs native} by analyzing next-token predictions across layers using logit lens. In Figure \ref{fig:romanization early}, panels \ref{fig:fr-ml} and \ref{fig:fr-ta} illustrate that for translations into native scripts, target language tokens begin to emerge from layer 40 onward. Conversely, in panels \ref{fig:fr-ml_translit} and \ref{fig:fr-ta_translit}, where the target language is in romanized script, target tokens appear 1–2 layers earlier. This pattern indicates that \textbf{when processing non-Roman script languages, the model forms internal representations of target tokens in earlier layers for romanized script compared to native script.} This trend is consistent across language pairs and models (c.f. Figures \ref{fig:romanization early gemma 2 9b it fr appendix}-\ref{fig:romanization early llama 2 13b fr appendix} in the Appendix). This suggests that romanization facilitates faster progression toward language-specific embeddings. 

 In Figure \ref{fig:romanization early}, in all four graphs consistent with \citet{wendler-etal-2024-llamas}, English representations emerge from the middle layers and persist until the final few layers, where the target language representations gradually take shape. It is important to note that native script curves (Figures \ref{fig:fr-ml}, \ref{fig:fr-ta})  exhibit steeper gradients than their romanized equivalents (Figures \ref{fig:fr-ml_translit}, \ref{fig:fr-ta_translit}).

\paragraph{Discussion.} In our investigation of romanized representations in the latent layers, we conclusively identified romanized tokens in the last 6–7 layers of an LLM across various multilingual text completion tasks. Based on previous works in this field \cite{wendler-etal-2024-llamas,zhao2024large}, in an English-centric decoder only LLM this region corresponds to the transition from an English-centric language-agnostic concept space to a language-specific space where the idea conceived in the concept space is expressed in the target language. Our findings suggest that \textbf{romanization serves as a bridge between the concept space and the language-specific region for non-Roman script languages}, an observation strongly supported by our analysis of six diverse writing systems. Romanization acting as a bridge could explain why romanization based script barrier breaking methods like \citet{liu-etal-2024-translico} and \citet{xhelili-etal-2024-breaking} work. Notably, we do not observe Latent Romanization in Chinese, likely due to its logographic script and relatively high-resource status.

\section{Conclusion} 
Our findings show that LLMs implicitly use Romanization as a bridge for non-Roman scripts, exhibiting \emph{Latent Romanization} in intermediate layers before switching to native scripts. Layerwise analyses reveal that semantic concepts are encoded similarly across native and Romanized inputs, indicating a shared internal representation. Moreover, when translating into a Romanized script, target words emerge earlier, highlighting Romanization as a structural link between language-agnostic concepts and language-specific output. While our study reveals initial insights into Latent Romanization, future work could focus on applying these findings to develop training strategies that enhance performance across diverse linguistic communities.

\section{Limitations}

The handling of multilingual text by large language models (LLMs) remains an active area of research. Although evidence suggests that LLMs process English representations within a language-agnostic space, the specific mechanisms by which these models adjust their interactions over different timesteps during token generation are still not fully understood. In our study, we observe that romanized representations become increasingly prominent in the hidden layers as token generation progresses from the first to the final token. This trend suggests that latent romanization may help the model mitigate differences in token fertility—that is, the average number of tokens required to represent a word—between the output language and its primary latent language, English. This effect appears especially for non-Roman script languages, with high token fertility. However, further research is needed to confirm and generalize these observations.

The interpretability of non-Roman scripts at latent layers is limited when models employ tokenization schemes that split non-Roman characters into multiple bytes, complicating logit lens analysis. Extending this work to models with alternative tokenization methods would offer a more complete understanding of multilingual capabilities and representations.

This work identifies but does not explain the selective occurrence and varying intensity of latent romanization across languages—questions that merit dedicated future investigation.

\section{Ethics Statement}

Through this work, our aim is to democratize access to LLMs and address the issue of limited data availability for low-resource languages. We emphasize that it is not our intention to diminish the value or significance of the native scripts of the languages included in this study.

The code and datasets created in this work will be made available under permissible licenses. Generative AI systems were only used for assistance purely with the language of the paper, e.g., paraphrasing, spell-check, polishing the author’s original content, and for writing boiler-plate code.

\section*{Acknowledgments}
We would like to thank EkStep Foundation and Nilekani Philanthropies for their generous grant towards research at AI4Bharat. We have adopted \citet{nina-rimsky} to interpret the LLMs. We have utilized the experimental setups and datasets provided by \citet{wendler-etal-2024-llamas} and \citet{dumas2024separating}.

\bibliography{anthology_reduced,custom}

\appendix

\section{Transformer's Forward pass: Detailed}
\label{sec:detailed_transformers_forward_pass}

For an input sequence $x_1, \ldots, x_n \in V$, where $n$ is the sequence length, the initial latents $h^{(0)}_1, \ldots, h^{(0)}_n \in \mathbb{R}^d$ are obtained from a learned embedding matrix. The update rule for the latent at position $i$ in layer $j$ is expressed as:
\[
    h^{(j)}_i = h^{(j-1)}_i + f_j(h^{(j-1)}_1, \ldots, h^{(j-1)}_i)
\]

The logit scores are computed as:
\[
    z_i = U h^{(k)}_i
\]

These are converted to probabilities via the softmax function:
\[
    P(x_{i+1} = t | x_1, \ldots, x_i) \propto \exp(z_{i,t})
\]

\section{Sample Prompts}
\label{sec:English_Translation_and_Romanization_of_the_Sample_Prompts}
A Hindi example, its English translation and transliteration for the translation, repetition and cloze task prompt designs mentioned in Section \ref{sec:methodology} are provided below.

\vspace{10pt}

\noindent\textbf{Hindi example.}

\vspace{0.25em}
\noindent\textit{Translation task.} A translation prompt from French to Hindi. 

 \vspace{10pt}

\setlength{\fboxsep}{5pt}
\setlength{\fboxrule}{1pt}
\ovalbox{
\begin{tabular}{l l}
\textbf{Français:} ``poisson" & \textbf{\dn Eh\306wdF}: ``{\dn mClF}$"$ \\
\textbf{Français:} ``mangue" & \textbf{\dn Eh\306wdF}: ``{\dn aAm}$"$ \\
\textbf{Français:} ``frère" & \textbf{\dn Eh\306wdF}: ``{\dn BAI}$"$ \\

\textbf{Français:} ``odeur" & \textbf{\dn Eh\306wdF}:``{\dn g\2D}$"$ \\ 
\textbf{Français:} ``soleil" & \textbf{\dn Eh\306wdF}: ``{\dn \8{s}rj}$"$ \\
\textbf{Français:} ``fleur" & \textbf{\dn Eh\306wdF}: \\ 


\end{tabular}
}
\vspace{10pt} 

\noindent\textit{Repetition task.}  \\

 \vspace{10pt} 

\setlength{\fboxsep}{5pt}
\setlength{\fboxrule}{1pt}
\ovalbox{
\begin{tabular}{l l}
\textbf{\dn Eh\306wdF}: ``{\dn mClF}$"$ & \textbf{\dn Eh\306wdF}: ``{\dn mClF}$"$ \\ 
\textbf{\dn Eh\306wdF}: ``{\dn aAm}$"$ & \textbf{\dn Eh\306wdF}: ``{\dn aAm}$"$ \\
\textbf{\dn Eh\306wdF}: ``{\dn BAI}$"$ & \textbf{\dn Eh\306wdF}: ``{\dn BAI}$"$ \\
\textbf{\dn Eh\306wdF}: ``{\dn g\2D}$"$ & \textbf{\dn Eh\306wdF}: ``{\dn g\2D}$"$ \\
\textbf{\dn Eh\306wdF}: ``{\dn \8{s}rj}$"$ & \textbf{\dn Eh\306wdF}: ``{\dn \8{s}rj}$"$ \\
\textbf{\dn Eh\306wdF}: ``\dn {\dn \8{P}l}$"$ & \textbf{\dn Eh\306wdF}:  \\
\end{tabular}
}

\vspace{10pt} 

\noindent\textit{Cloze task.}.

\vspace{10pt} 

\setlength{\fboxsep}{5pt}
\setlength{\fboxrule}{1pt}

\fbox{%
    \begin{tabular}{@{}p{0.9\linewidth}@{}}
    
    {\dn ek $``\_\_\_"$ khAEnyA\1 pxn\? k\? Ele upyog EkyA jAtA h\4. u\381wr{\rs :\re}$``$\7{p}-tk$"$}\\
    {\dn \7{P}VbA\<l aOr bA-k\?VbA\<l j\4s\? K\?l K\?ln\? k\? Ele $``\_\_\_"$ kA upyog EkyA jAtA h\4. u\381wr{\rs :\re} $``$bA\<l$"$} \\
    {\dn ek $``\_\_\_"$ a?sr uphAr k\? !p m\?{\qva} EdyA jAtA h\4 aOr yh bgFco{\qva} m\?{\qva} pAyA jA sktA h\4. u\381wr{\rs :\re}} \\
    \end{tabular}%
}


\vspace{10pt}

\noindent\textbf{English Translation.}

\vspace{5pt}

\noindent\textit{Translation task.}

\vspace{10pt} 

\setlength{\fboxsep}{5pt}
\setlength{\fboxrule}{1pt}
\ovalbox{
\begin{tabular}{l l}
\textbf{Français:} ``poisson" & \textbf{Hindi}: ``{fish}$"$ \\
\textbf{Français:} ``mangue" & \textbf{Hindi}: ``{mango}$"$ \\
\textbf{Français:} ``frère" & \textbf{Hindi}: ``{brother}$"$ \\

\textbf{Français:} ``odeur" & \textbf{Hindi}: ``{smell}$"$ \\ 
\textbf{Français:} ``soleil" & \textbf{Hindi}: ``{sun}$"$ \\ 
\textbf{Français:} ``fleur" & \textbf{Hindi}: \\ 
 
\end{tabular}
}
\vspace{10pt} 

\noindent\textit{Repetition task.}  \\

\setlength{\fboxsep}{5pt}
\setlength{\fboxrule}{1pt}
\ovalbox{
\begin{tabular}{l l}
\textbf{Hindi}: ``{fish}$"$ & \textbf{Hindi}: ``{fish}$"$ \\ 
\textbf{Hindi}: ``{mango}$"$ & \textbf{Hindi}: ``{mango}$"$ \\
\textbf{Hindi}: ``{brother}$"$ & \textbf{Hindi}: ``{brother}$"$ \\
\textbf{Hindi}: ``{smell}$"$ & \textbf{Hindi}: ``{smell}$"$ \\
\textbf{Hindi}: ``{sun}$"$ & \textbf{Hindi}: ``{sun}$"$ \\
\textbf{Hindi}: ``{flower}$"$ & \textbf{Hindi}:  \\
\end{tabular}
}
\vspace{10pt} 

\noindent\textit{Cloze task.}  \\

\vspace{10pt} 

\setlength{\fboxsep}{5pt}
\setlength{\fboxrule}{1pt}

\fbox{%
    \begin{tabular}{@{}p{0.9\linewidth}@{}}
    A ``\_\_\_" is used to play sports like soccer and basketball. Answer: ``ball" \\ 
    A ``\_\_\_" is used for reading stories. Answer: ``book" \\ 
    A ``\_\_\_" is often given as a gift and can be found in gardens. Answer:  \\ 
    \end{tabular}%
}
 \vspace{10pt} 

\noindent\textbf{English Transliteration.}

\vspace{5pt}

\noindent\textit{Translation task.}

\vspace{10pt} 

\setlength{\fboxsep}{5pt}
\setlength{\fboxrule}{1pt}
\ovalbox{
\begin{tabular}{l l}
\textbf{Français:} ``poisson" & \textbf{Hindi}: ``{machhalee}$"$ \\
\textbf{Français:} ``mangue" & \textbf{Hindi}: ``{aam}$"$ \\
\textbf{Français:} ``frère" & \textbf{Hindi}: ``{bhaee}$"$ \\

\textbf{Français:} ``odeur" & \textbf{Hindi}: ``{gandh}$"$ \\ 
\textbf{Français:} ``soleil" & \textbf{Hindi}: ``{sooraj}$"$ \\ 
\textbf{Français:} ``fleur" & \textbf{Hindi}: \\ 
 
\end{tabular}
}
\vspace{10pt} 

\noindent\textit{Repetition task.}  \\

\setlength{\fboxsep}{5pt}
\setlength{\fboxrule}{1pt}
\ovalbox{
\begin{tabular}{l l}
\textbf{Hindi}: ``{machhalee}$"$ & \textbf{Hindi}: ``{machhalee}$"$ \\ 
\textbf{Hindi}: ``{aam}$"$ & \textbf{Hindi}: ``{aam}$"$ \\
\textbf{Hindi}: ``{bhaee}$"$ & \textbf{Hindi}: ``{bhaee}$"$ \\
\textbf{Hindi}: ``{gandh}$"$ & \textbf{Hindi}: ``{gandh}$"$ \\
\textbf{Hindi}: ``{sooraj}$"$ & \textbf{Hindi}: ``{sooraj}$"$ \\
\textbf{Hindi}: ``{phool}$"$ & \textbf{Hindi}:  \\
\end{tabular}
}
\vspace{10pt} 

\noindent\textit{Cloze task.}  \\

\vspace{10pt} 

\setlength{\fboxsep}{5pt}
\setlength{\fboxrule}{1pt}

\fbox{%
    \begin{tabular}{@{}p{0.9\linewidth}@{}}
    Phutball aur baasketball jaise khel khelane ke lie ``\_\_\_" ka upayog kiya jaata hai. Uttar: ``ball" \\ 
    Ek ``\_\_\_" kahaaniyaan padhane ke lie upayog kiya jaata hai. Uttar: ``ball"
    Ek ``\_\_\_" aksar upahaar ke roop mein diya jaata hai aur yah bageechon mein paaya ja sakata hai. Uttar: \\ 
    \end{tabular}%
}
 \vspace{10pt}

 \section{Latent romanization: Tokenization scheme for the romanized word}
 \label{sec:latent_romanization_tokenization_scheme}
  Mathematically, we track the following romanized tokens for a given romanized word $w$ of length $n$:\ 

\vspace{5pt} 
\setlength{\fboxsep}{5pt}
\setlength{\fboxrule}{1pt}

\fbox{%
    \begin{tabular}{@{}p{0.9\linewidth}@{}}
First timestep: \( \{ w[0:i] \mid 1 \leq i \leq n \} \cup \{ \text{"\_ "} + w[0:i] \mid 1 \leq i \leq n \} \), "\_ " represents single leading space and $w[0:i]$ represents prefixes of $w$ \\
Intermediate timesteps: \( \{w[i:j] \mid 0 \leq i < j \leq n\} \) , $w[i:j]$ represents sub-strings of $w$\\
Final timestep: \( \{w[i:n] \mid 0 \leq i < n\} \), , $w[i:n]$ represents suffixes of $w$

    \end{tabular}%
}
\vspace{5pt}

Similarly we construct token sets for the native script ($T_{\text{native}}$) and English ($T_{\text{english}}$) by including prefixes of the corresponding word, both with and without leading spaces, for all timesteps except the last. For the final timestep, we use suffixes of the corresponding word. We discard the sample if there is any overlap between romanized tokens ($T_{\text{romanized}}$) and either the native ($T_{\text{native}}$) or English tokens ($T_{\text{english}}$) as shown by the following condition: \\
\[ 
T_{\text{romanized}} \cap (T_{\text{native}} \cup T_{\text{English}}) = \emptyset 
\]

Lets take an example with a prompt translating $``rope"$ from French to Hindi and derive romanized, English and native tokens for its first token generation timestep. The Hindi translation of  $``rope"$ is {\dn r-sF} and its romanized form is $``rassi"$. So the romanized word tokens would be $T_{\text{romanized}}$ = $``r"$, $``ra"$, $``ras"$, $``rass"$, $``rassi"$, $``\_r"$, $``\_ra"$, $``\_ras"$, $``\_rass"$ and $``\_rassi"$. The corresponding English word tokens would be $T_{\text{English}}$ =  $``r"$, $``ro"$, $``rop"$, $``rope"$, $``\_r"$, $``\_ro"$, $``\_rop"$, and $``\_rope"$. The corresponding Hindi word tokens would be $T_{\text{native}}$ = {\dn r}, {\dn rs}, {\dn r-s}, {\dn r-sF}, {\dn \_r}, {\dn \_rs}, {\dn \_r-s}, {\dn \_r-sF}. Here $ T_{\text{romanized}} \cap (T_{\text{native}} \cup T_{\text{English}}) =  \{r, \_r\}$  which is not null. As such we will exclude this example translating rope to Hindi from the dataset to analyze \textit{Latent Romanization}.

\vspace{5pt}

\section{Latent Fraction}
\label{sec:latent_fraction}
Formally, we compute the latent fraction as follows:

\vspace{5pt} 
\setlength{\fboxsep}{5pt}
\setlength{\fboxrule}{1pt}

\fbox{%
    \begin{tabular}{@{}p{0.9\linewidth}@{}}
    For layer $l$, timestep $t$, sample $i$ and set of corresponding romanized tokens $R$ :

1. \textbf{Latent romanization condition:} \[
r_{l,t}^{(i)} =
\begin{cases} 
      1, & \text{if } \max\limits_{r \in R} P(x_{t} = r \mid l,t) > 0.1 \\
      0, & \text{otherwise}
\end{cases}
\]

2. \textbf{Latent fraction for a layer $\ell$:} \\

       \[
\text{L.F}(l) = \frac{1}{N} \sum_{i=1}^N \frac{1}{T} \sum_{t=1}^T r_{l,t}^{(i)}
\]

where $N$ is the number of samples, $T$ is the number of generation timesteps and  $P(x_t = r|l,t)$ is the probability of generating token $r$ at timestep $t$ and layer $\ell$.
    \end{tabular}%
}

\section{Computing Language probabilities - For translation towards native script vs translation towards romanized script task}
\label{sec:Computing_Language_probabilities}
To compute language probabilities, we search the LLM's vocabulary for all tokens that could be the first token of the correct ouput word(s) in the respective language. We search the models vocabulary for all prefixes of the word(s) with and without leading space. For a language $\ell$ with corresponding output word $w_1$ and its synonyms $w_2, w_3, \ldots$, we define:
\[
P(\text{lang} = \ell) = \sum_{t_\ell \in W_\text{prefix}} P(x_{n+1} = t_\ell)
\]
where $W_\text{prefix}$ is the set of all prefixes of output word $w_1$ and its synonyms $w_2, w_3, \ldots$, including versions with and without leading spaces. For example to get probability for english when the output word is ``fast$"$ and its synonym is ``swift$"$, then $P(\text{lang} = \text{EN}) = P(x_{n+1} = \text{``f$"$}) + P(x_{n+1} = \text{``fa$"$}) + P(x_{n+1} = \text{``fas$"$}) + P(x_{n+1} = \text{``fast$"$}) + P(x_{n+1} = \text{``\_f$"$}) + P(x{n+1} = \text{``\_fa$"$}) + P(x{n+1} = \text{``\_fas$"$}) + P(x{n+1} = \text{``\_fast$"$}) + P(x{n+1} = \text{``s$"$}) + P(x_{n+1} = \text{``sw$"$}) + P(x_{n+1} = \text{``swi$"$}) + P(x_{n+1} = \text{``swif$"$}) + P(x_{n+1} = \text{``swift$"$}) + P(x_{n+1} = \text{``\_s$"$}) + P(x{n+1} = \text{``\_sw$"$}) + P(x{n+1} = \text{``\_swi$"$}) + P(x{n+1} = \text{``\_swif$"$}) + P(x{n+1} = \text{``\_swift$"$})$ (all the token-level prefixes of ``fast$"$, ``\_fast$"$, ``swift$"$ and ``\_swift$"$). ``\_$"$ represents a single leading space.

\section{Romanization scheme: Indic languages}
\label{sec:romanization_scheme_indic_languages}
 We have taken into consideration two romanization schemes for Indic languages: (a) ITRANS scheme from IndicNLP library \cite{Kunchukuttan-2020-IndicNLP}  and (b) IndicXlit scheme \cite{madhani-etal-2023-aksharantar}. Based on our initial experiments, we observed that the IndicXlit scheme produces better romanization than ITRANS scheme. Thus for romanization we have used the IndicXlit scheme \cite{madhani-etal-2023-aksharantar}. It generates romanization as is commonly used by native speakers and is trained on parallel transliteration corpora.

\section{Computing Probabilities : Activation Patching Experiment}
\label{sec:Computing_Probability_Activation_Patching_Experiment}
Probability for a concept $C$ in language $\ell$ can be formulated as :

\[
P(C^\ell) = \sum_{t_\ell \in W_\text{prefix}} P(x_{n+1} = t_\ell)
\]
where $W_\text{prefix}$ is the set of all prefixes of output word $w(C^\ell)$ and its synonyms (note that a word's tokens are its prefixes).

We keep source concept $C_S$ and target concept $C_T$ distinct to avoid ambiguity when both are expressed in the same target language $l^T_{out}$.

Cases of token overlap between $w(C^{\ell_T}_S)$ i.e. word representing source concept in  target language  and $w(C^{\ell_T}_T)$ i.e. word representing target concept in  target language and their synonyms 
are excluded. Token overlap would cause ambiguity. Therefore in the final dataset,

\[
T(w(C_S^{\ell_T})) \cap T(w(C_T^{\ell_T})) = \emptyset
\]
Where $T(w)$ represents all the prefixes of $w$ and its synonyms.

\section{Latent Romanization Qualitative Analysis}
\label{sec:logit_lens_qualitative_analysis_appendix}
\begin{figure*}[h!]
  \includegraphics[width= \textwidth]{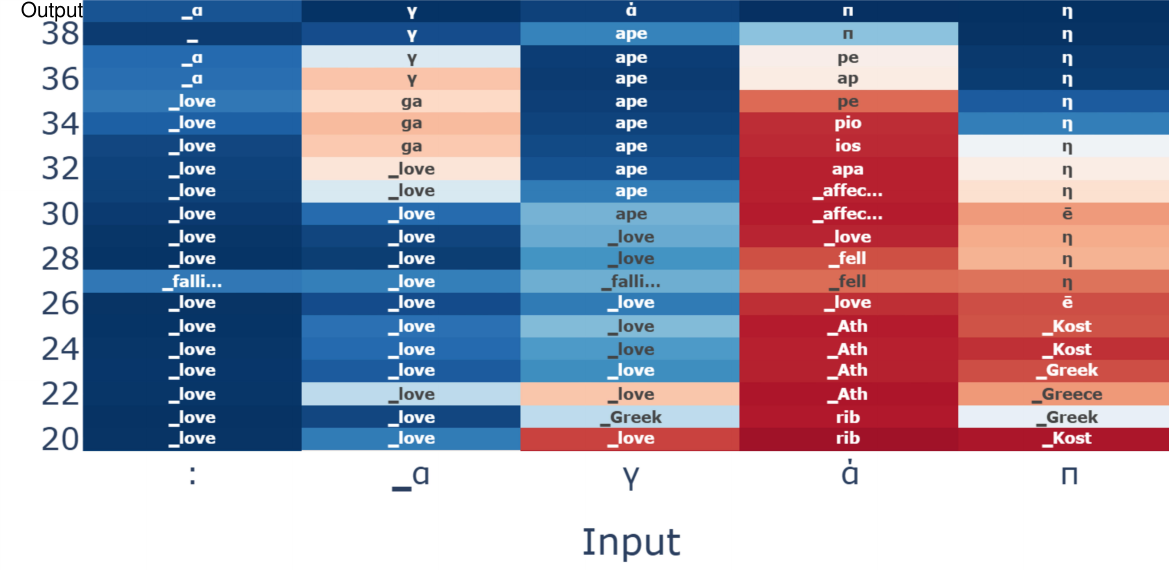}
    \caption{\textbf{Logit lens illustration.} We input  Llama-2 13b model with a  prompt 
    translating `love' from French to Greek. We visualize the output \selectlanguage{greek}
\textgreek{(αγάπη}  
\selectlanguage{english}, `agape' is the romanized form ) taking shape using logit lens producing a next-token distribution for each position (x-axis) and layers 20 and above (y-axis).Interestingly in the middle to top layers (20 - 29) we could observe romanized subwords of the Greek word (\selectlanguage{greek}
\textgreek{γ}  
\selectlanguage{english} - ga ; \selectlanguage{greek}
\textgreek{άπη}  
\selectlanguage{english} - ape ;
\selectlanguage{greek}
\textgreek{πη}  
\selectlanguage{english} -pe ;   \selectlanguage{greek}
\textgreek{η}  
\selectlanguage{english} - e) before they are represented in their native script. Color represents entropy of next-token generation from low (blue) to high (red). Plotting tool: \cite{belrose2023eliciting}. 
}
    \label{fig:logit_lens_love_greek}
\end{figure*}

\begin{figure*}[h!]
  \includegraphics[width= \textwidth]{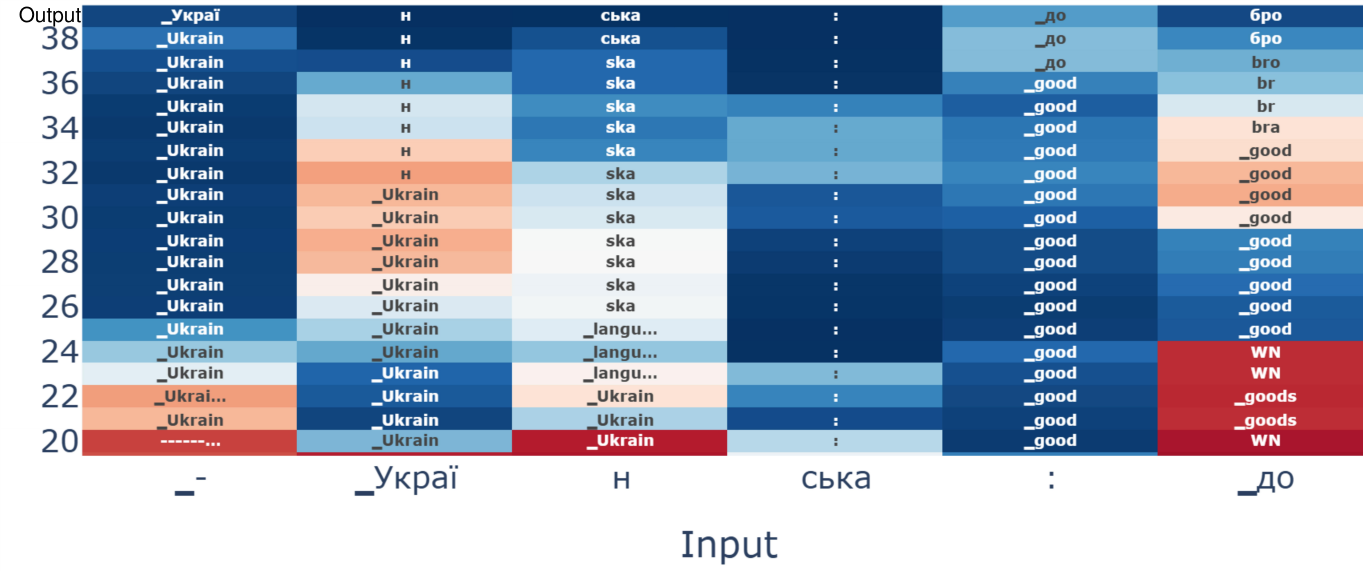}
    \caption{\textbf{Logit lens illustration.} We input  Llama-2 13b model with a  prompt 
    translating `good' from French to Ukrainian. We visualize the output (\selectlanguage{ukrainian}
 добро
\selectlanguage{english}, `dobro' is the romanized form ) taking shape using logit lens producing a next-token distribution for each position (x-axis) and layers 20 and above (y-axis).'\selectlanguage{ukrainian}
 Українська
\selectlanguage{english}' (romanized as 'Ukrayinska') is the Ukrainian word for 'Ukrainian'.Interestingly in the middle to top layers (20 - 29) we could observe romanized subwords of the Ukrainian words (\selectlanguage{ukrainian}
бро
\selectlanguage{english} - bro ; \selectlanguage{ukrainian}
ська
\selectlanguage{english}  - ska ) before they are represented in their native script. Color represents entropy of next-token generation from low (blue) to high (red). Plotting tool: \cite{belrose2023eliciting}. 
}
    \label{fig:logit_lens_ukrainian_good}
\end{figure*}

\begin{figure*}[h!]
  \includegraphics[width= \textwidth]{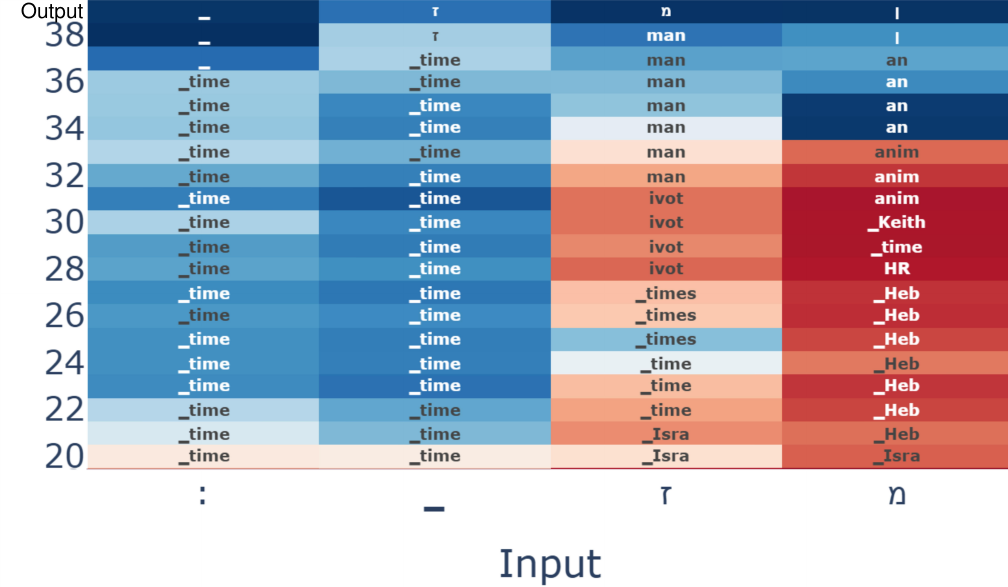}
    \caption{\textbf{Logit lens illustration.} We input  Llama-2 13b model with a  prompt 
    translating `time' from French to Hebrew. We visualize the output ({\cjhebrew{Nmz}}, `zman' is the romanized form ) taking shape using logit lens producing a next-token distribution for each position (x-axis) and layers 20 and above (y-axis).Interestingly in the middle to top layers (20 - 29) we could observe romanized subwords of the Hebrew word ({\cjhebrew{Nm}} - man ; {\cjhebrew{N}} - an ). Color represents entropy of next-token generation from low (blue) to high (red). Plotting tool: \cite{belrose2023eliciting}. 
}
    \label{fig:logit_lens_hebrew_time}
\end{figure*}

\begin{figure*}[h!]
  \includegraphics[width= \textwidth]{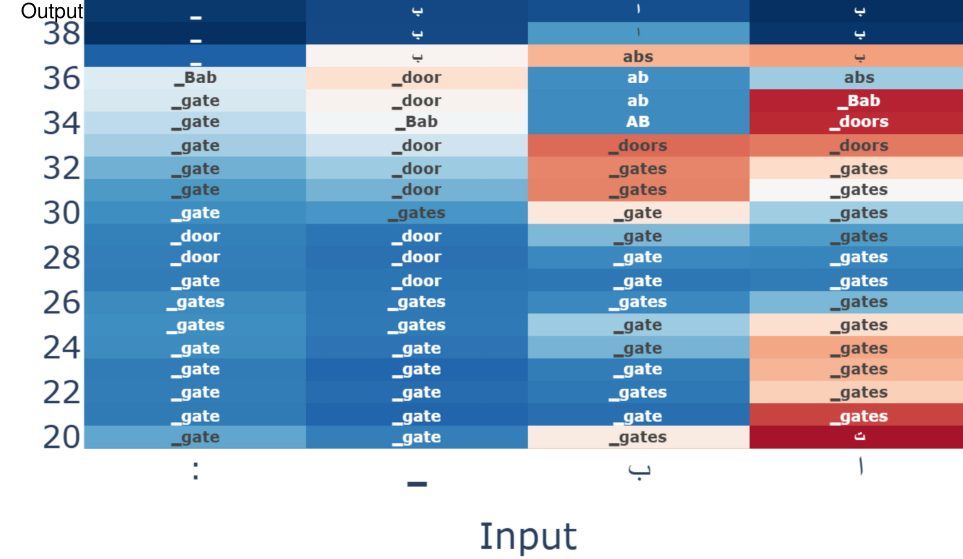}
    \caption{\textbf{Logit lens illustration.} We input  Llama-2 13b model with a  prompt 
    translating `door' from French to Arabic. We visualize the output (\raisebox{-1ex}{\includegraphics[height=3ex]{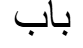}} , `bab' is the romanized form ) taking shape using logit lens producing a next-token distribution for each position (x-axis) and layers 20 and above (y-axis).Interestingly in the middle to top layers (20 - 29) we could observe romanized subwords of the Arabic word (\raisebox{-1ex}{\includegraphics[height=3ex]{latex/fig/appendix_arab_words/arabic_bab.png}}- bab;\raisebox{-1ex}{\includegraphics[height=3ex]{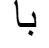}}- ab) before they are represented in their native script. Color represents entropy of next-token generation from low (blue) to high (red). Plotting tool: \cite{belrose2023eliciting}. 
}
    \label{fig:logit_lens_arabic_door}
\end{figure*}

\begin{figure*}[h!]
  \includegraphics[width= \textwidth]{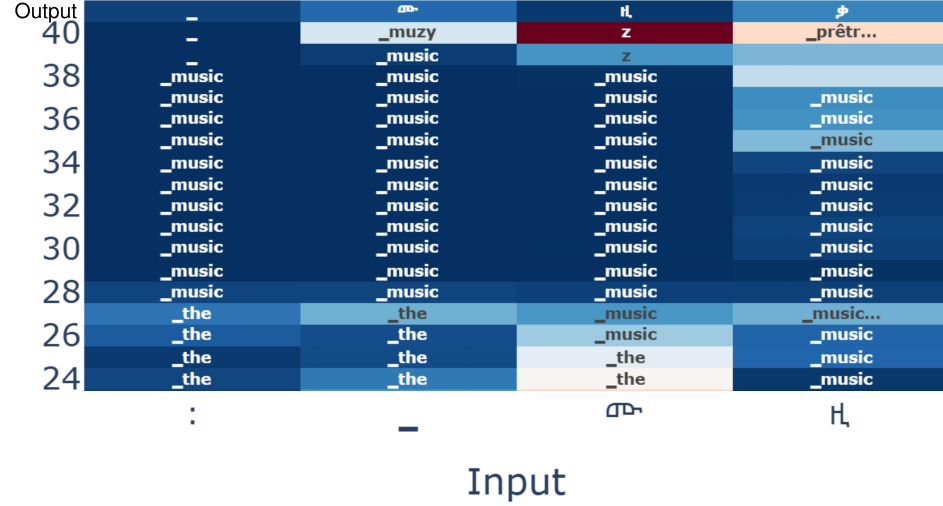}
    \caption{\textbf{Logit lens illustration.} We input  Gemma-2 9b IT model with a  prompt 
    translating `music' from French to Amharic. We visualize the output (\raisebox{-0.5ex}{\includegraphics[height=2ex]{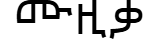}}, `muzika' is the romanized form ) taking shape using logit lens producing a next-token distribution for each position (x-axis) and layers 24 and above (y-axis).Interestingly in the middle to top layers  we could observe romanized subwords of the Amharic word (\raisebox{-0.5ex}{\includegraphics[height=2ex]{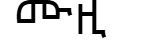}}- muzy; \raisebox{-0.5ex}{\includegraphics[height=2ex]{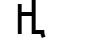}}- z) before they are represented in their native script. Color represents entropy of next-token generation from low (blue) to high (red). Plotting tool: \cite{belrose2023eliciting}. 
}
    \label{fig:logit_lens_amharic_music}
\end{figure*}

We list qualitative logit lens analysis for Greek, Ukrainian, Hebrew, Arabic and Amharic (see Figures \ref{fig:logit_lens_love_greek} to \ref{fig:logit_lens_amharic_music}).

\paragraph{Languages.}Greek is part of the Hellenic branch of the Indo-European language family and is written using the Greek alphabet. Ukrainian belongs to the East Slavic group of the Indo-European family and employs the Ukrainian alphabet, a variant of the Cyrillic script. Amharic is a South Ethio-Semitic language within the Afroasiatic family and is written using the Ge'ez script, an abugida where each character represents a consonant-vowel combination. Hebrew is a Northwest Semitic language within the Afroasiatic family and is written using the Hebrew alphabet, an abjad script originating from the Aramaic alphabet. Arabic is a Central Semitic language, also part of the Afroasiatic family, and utilizes the Arabic script, another abjad that evolved from the Nabataean alphabet. Notably, both Hebrew and Arabic scripts are written from right to left.

\section{Latent Romanization Quantitative Analysis: Additional examples}
\begin{figure*}[htbp]
    \centering
    \begin{subfigure}{0.3\textwidth} 
        \centering
        \includegraphics[width=\textwidth]{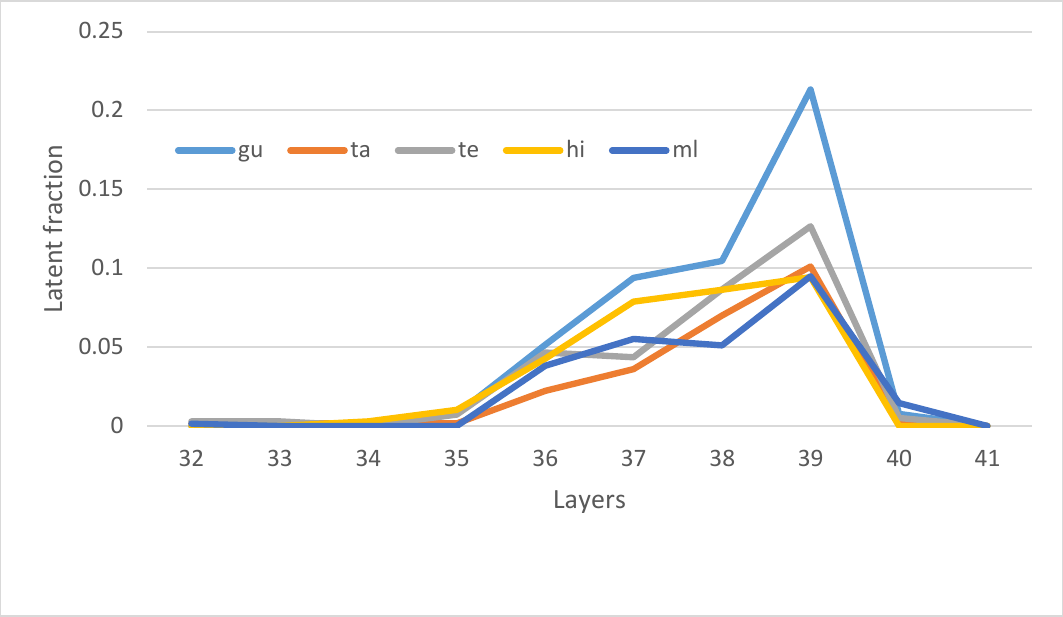} 
        \caption{Layerwise distribution of romanized tokens averaged across output token generation steps and samples }
        \label{fig:first_token_layerwise_lat_rom_gemma_9b_rep_all_tokens}
    \end{subfigure}
    \hfill
    \begin{subfigure}{0.3\textwidth} 
        \centering
        \includegraphics[width=\textwidth]{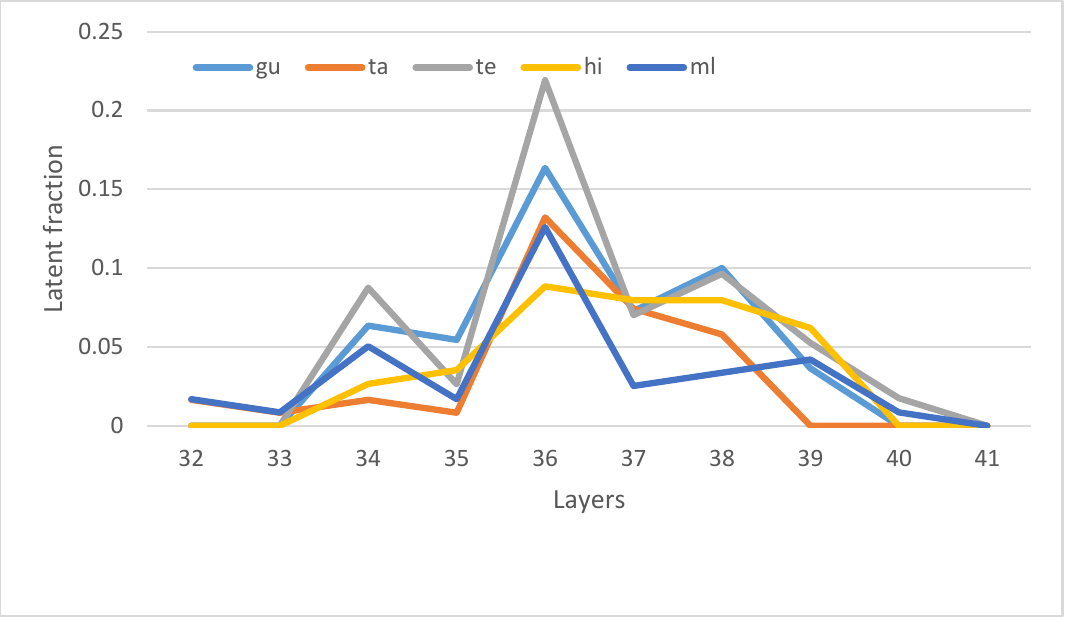} 
        \caption{Layerwise distribution of romanized tokens in the first token generation step averaged across samples }
        \label{fig:first_token_layerwise_lat_rom_gemma_9b_it_rep_first}
    \end{subfigure}
    \hfill
    \begin{subfigure}{0.3\textwidth} 
        \centering
        \includegraphics[width=\textwidth]{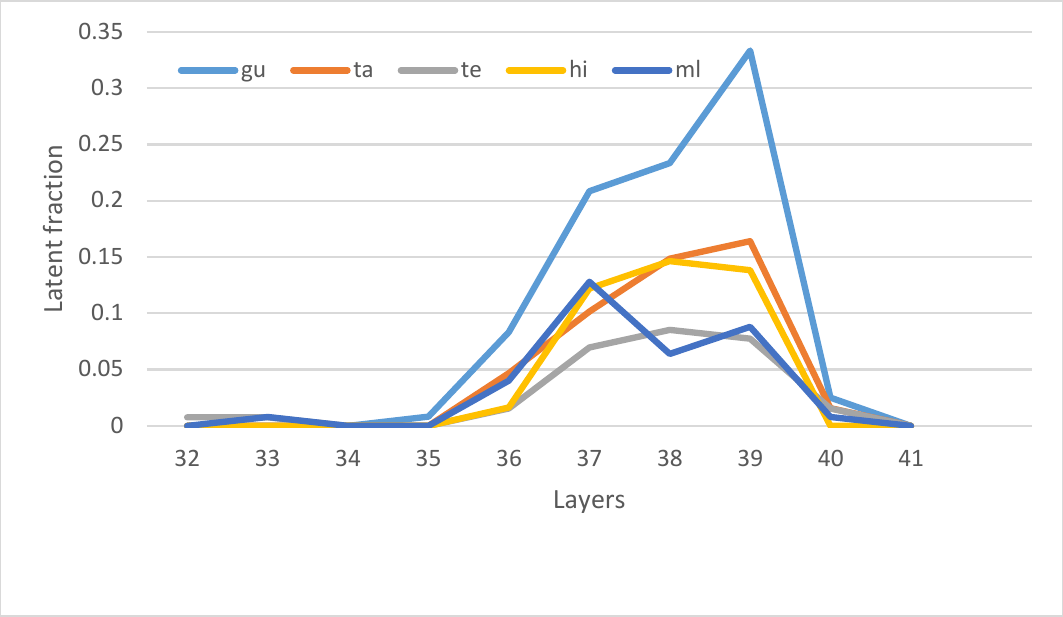} 
        \caption{Layerwise distribution of romanized tokens in the last token generation step averaged across samples}

    \end{subfigure}
     \hfill
     \caption{\textbf{Distribution of Romanized Tokens Across Model Layers: Analysis of First, Last, and All Generation Timesteps.} This distribution is plotted across the last 10 layers of Gemma-2 9b IT model for \textbf{repetition task} and is averaged across 100+ samples. \textit{X}-axes represents layer index, \textit{y}-axes represents latent fraction i.e. the instances where romanized tokens occur with a probability > 0.1 averaged over samples for a specific layer.  We plot the distributions for Gujarati (gu), Tamil (ta), Telugu (te), Hindi (hi) and Malayalam (ml).}
      
\label{fig:latent_romanization gemma 2 9b it repetition task}
\end{figure*}

\begin{figure*}[htbp]
    \centering
    \begin{subfigure}{0.3\textwidth} 
        \centering
        \includegraphics[width=\textwidth]{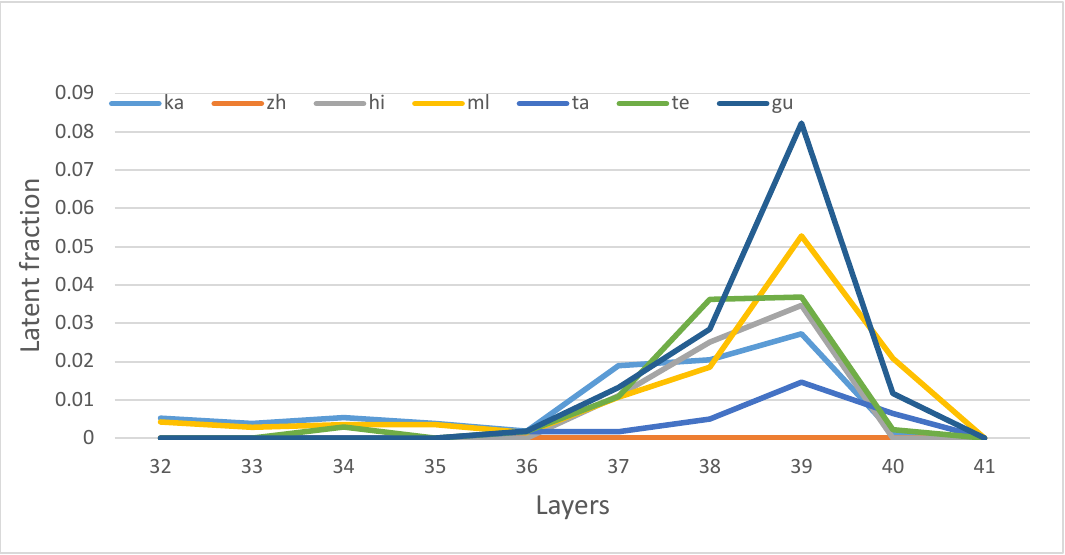} 
        \caption{Layerwise distribution of romanized tokens averaged across output token generation steps and samples }
        
    \end{subfigure}
    \hfill
    \begin{subfigure}{0.3\textwidth} 
        \centering
        \includegraphics[width=\textwidth]{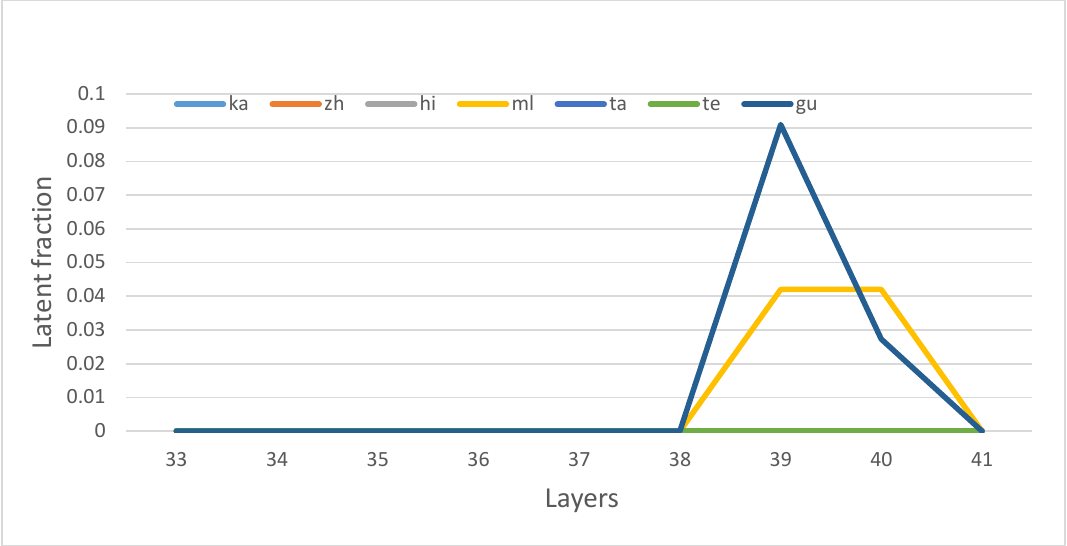} 
        \caption{Layerwise distribution of romanized tokens in the first token generation step averaged across samples }
        
    \end{subfigure}
    \hfill
    \begin{subfigure}{0.3\textwidth} 
        \centering
        \includegraphics[width=\textwidth]{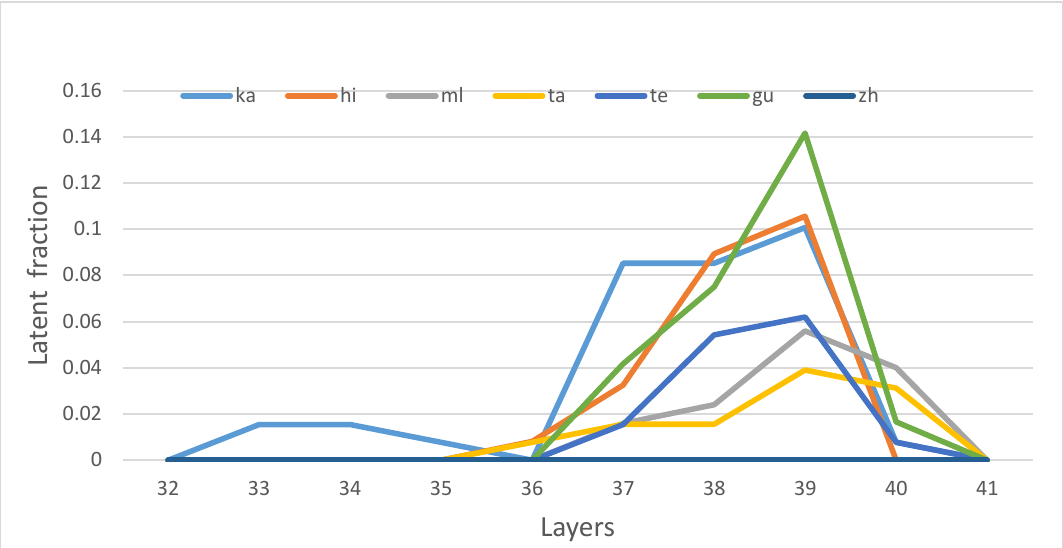} 
        \caption{Layerwise distribution of romanized tokens in the last token generation step averaged across samples}

    \end{subfigure}
     \hfill
     \caption{\textbf{Distribution of Romanized Tokens Across Model Layers: Analysis of First, Last, and All Generation Timesteps.} This distribution is plotted across the last 10 layers of Gemma-2 9b IT model for \textbf{cloze task} and is averaged across 100+ samples. \textit{X}-axes represents layer index, \textit{y}-axes represents latent fraction i.e. the instances where romanized tokens occur with a probability > 0.1 averaged over samples for a specific layer.  We plot the distributions for Gujarati (gu), Tamil (ta), Telugu (te), Hindi (hi) and Malayalam (ml), Georgian (ka) and Chinese (zh).}
      
\label{fig:latent_romanization gemma 2 9b it cloze task}
\end{figure*}

\begin{figure*}[htbp]
    \centering
    \begin{tabular}{@{}c@{\hspace{1em}}c@{\hspace{1em}}c@{}}
        \multicolumn{1}{c}{Gemma 2 9b} &
        \multicolumn{1}{c}{Llama 2 13b} &
        \multicolumn{1}{c}{Llama 2 7b} \\[1ex]

        \includegraphics[width=0.3\textwidth]{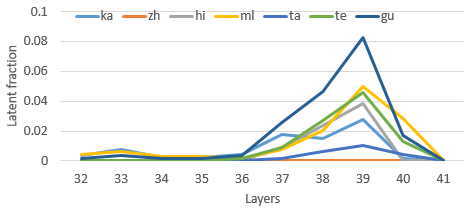} &
        \includegraphics[width=0.3\textwidth]{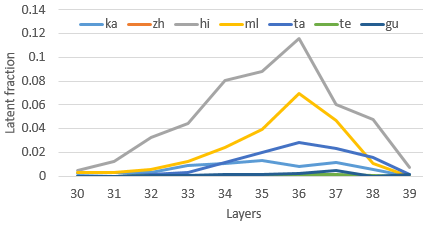} &
        \includegraphics[width=0.3\textwidth]{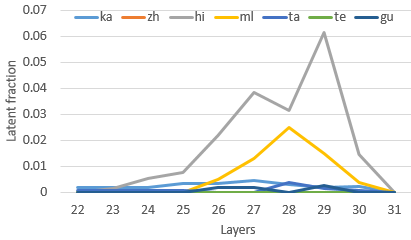} \\
    \end{tabular}
    
    Translation
    
    \vspace{2ex} 
    
    \begin{tabular}{ccc}
        \includegraphics[width=0.3\textwidth]{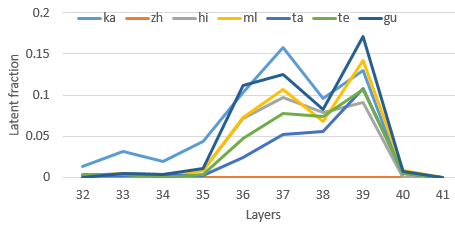} &
        \includegraphics[width=0.3\textwidth]{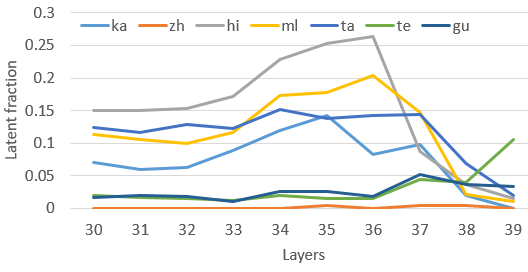} &
        \includegraphics[width=0.3\textwidth]{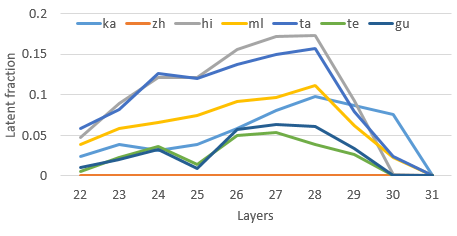} \\
    \end{tabular}
    
    Repetition

    \vspace{2ex} 
    
    \begin{tabular}{ccc}
        \includegraphics[width=0.3\textwidth]{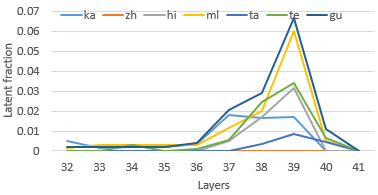} &
        \includegraphics[width=0.3\textwidth]{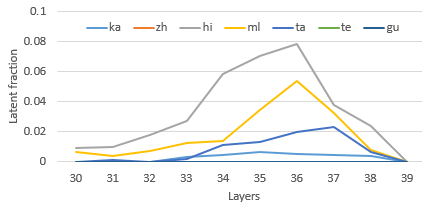} &
        \includegraphics[width=0.3\textwidth]{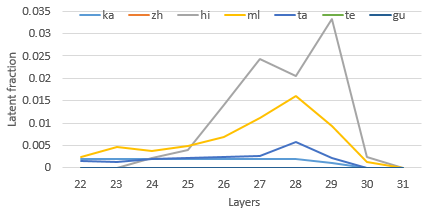} \\
    \end{tabular}
    
    Cloze
    
    \caption{\textbf{Layerwise fractional distribution of romanized tokens across output token generation timesteps.} This distribution is plotted across the last 10 layers of Gemma 2 9b ,LLama 2 13b and Llama 2 7b models (columns) for (a) translation task from French, (b) Repetition task and is averaged across 100+ samples. \textit{X}-axes represents layer index, \textit{y}-axes represents latent fraction i.e. the fraction of timesteps where romanized tokens occur with a probability > 0.1 averaged over samples for a specific layer. We plot the distributions for Gujarati (gu), Tamil (ta), Telugu (te), Hindi (hi) and Malayalam (ml).}
    \label{fig:all_tokens_latent_fraction_appendix}
\end{figure*}

Quantitative Analysis of latent romanization for repetition task and cloze task with gemma 2 9b it model can be seen in Figures \ref{fig:latent_romanization gemma 2 9b it repetition task} and \ref{fig:latent_romanization gemma 2 9b it cloze task} respectively. Layerwise fractional distribution of romanized tokens across output token generation timesteps for translation, repetition and cloze task with Gemma 2 9b, Llama 2 7b, and Llama 2 13b models are present in Figure \ref{fig:all_tokens_latent_fraction_appendix}. 

\section{Comparing Translations Into Romanized vs. Native Script: Additional examples}
\begin{figure*}[htbp]
    \centering
    
    \begin{subfigure}{0.19\textwidth}
        \centering
        \includegraphics[width=\textwidth]{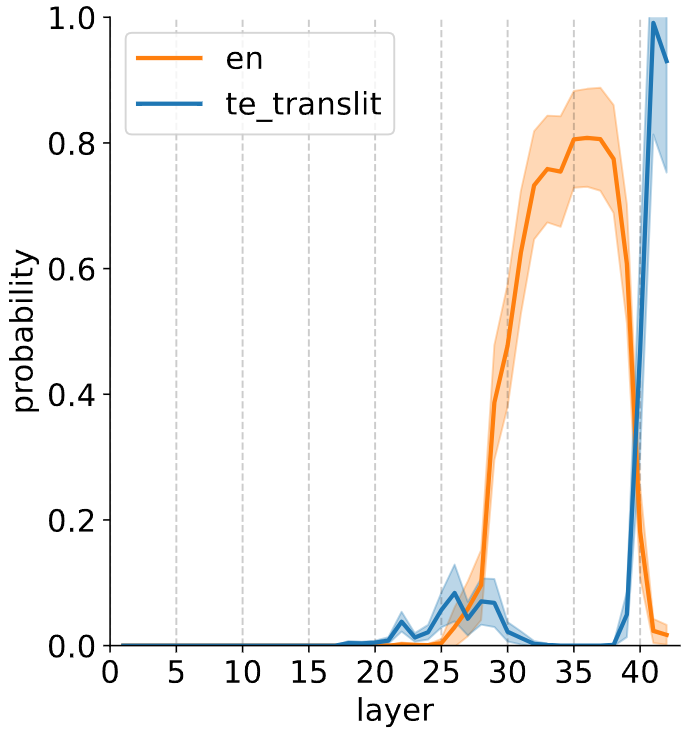}
        \caption{fr → te romanized}
        \label{fig:gemma 2 9b it fr-te_translit}
    \end{subfigure}
    \hfill
    \begin{subfigure}{0.19\textwidth}
        \centering
        \includegraphics[width=\textwidth]{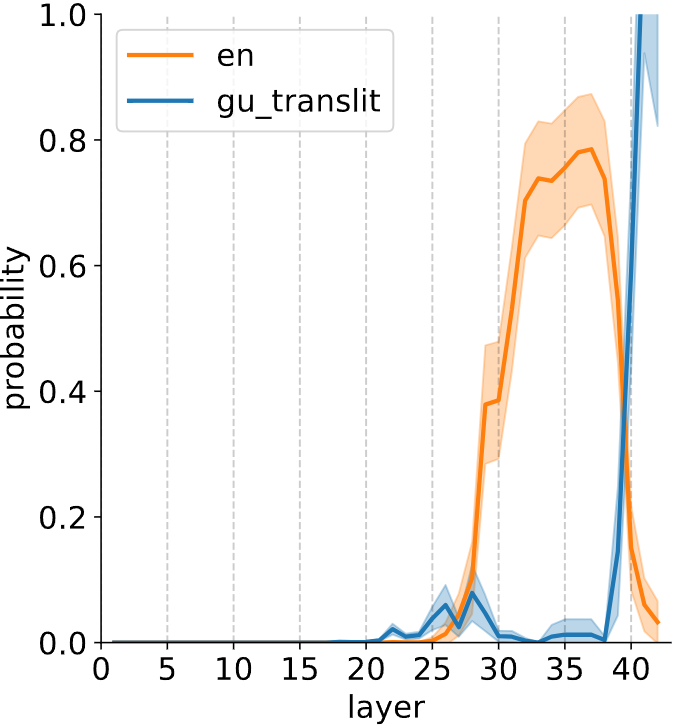}
        \caption{fr → gu romanized}
        \label{fig:gemma 2 9b it fr-gu_translit}
    \end{subfigure}
    \hfill
    \begin{subfigure}{0.19\textwidth}
        \centering
        \includegraphics[width=\textwidth]{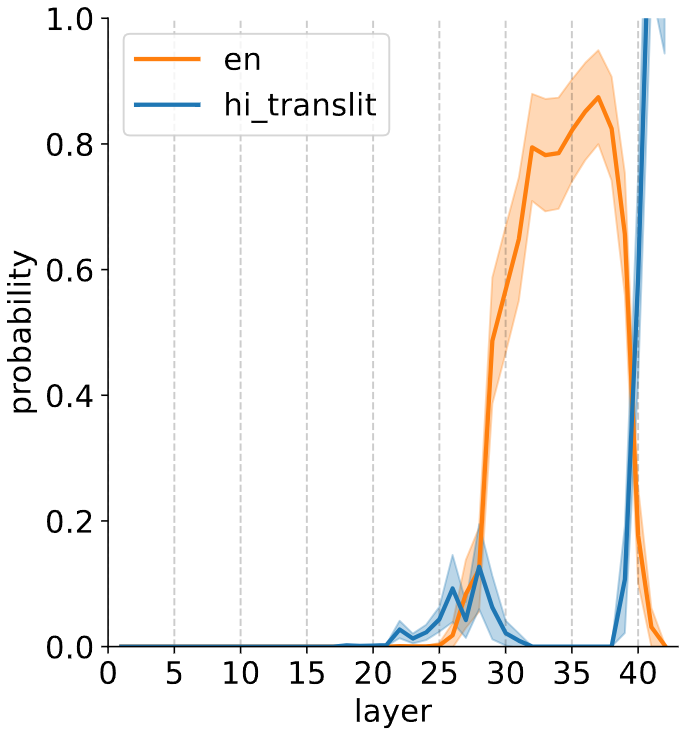}
        \caption{fr → hi romanized}
        \label{fig:gemma 2 9b it fr-hi_translit}
    \end{subfigure}
    
    \vspace{1em}  
    
    \begin{subfigure}{0.19\textwidth}
        \centering
        \includegraphics[width=\textwidth]{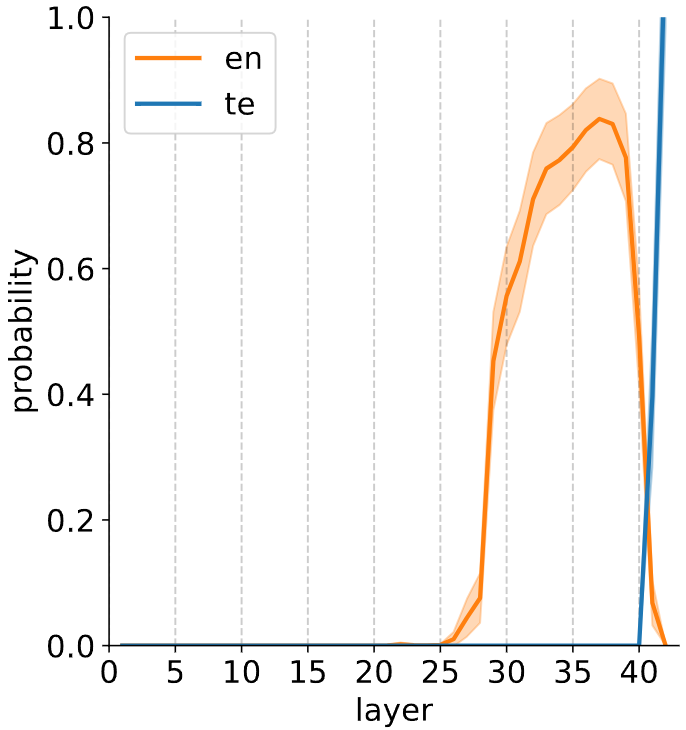}
        \caption{fr → te}
        \label{fig:gemma 2 9b it fr-te}
    \end{subfigure}
    \hfill
    \begin{subfigure}{0.19\textwidth}
        \centering
        \includegraphics[width=\textwidth]{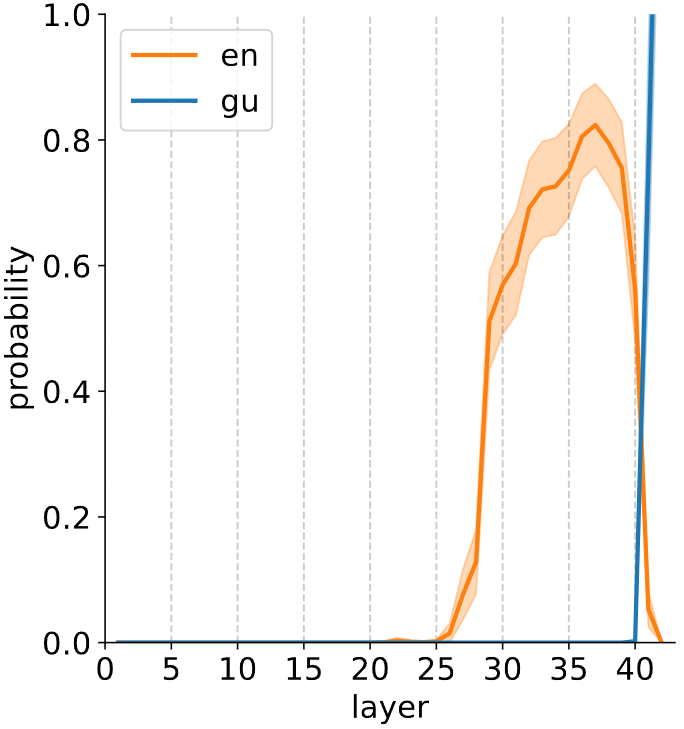}
        \caption{fr → gu}
        \label{fig:gemma 2 9b it fr-gu}
    \end{subfigure}
    \hfill
    \begin{subfigure}{0.19\textwidth}
        \centering
        \includegraphics[width=\textwidth]{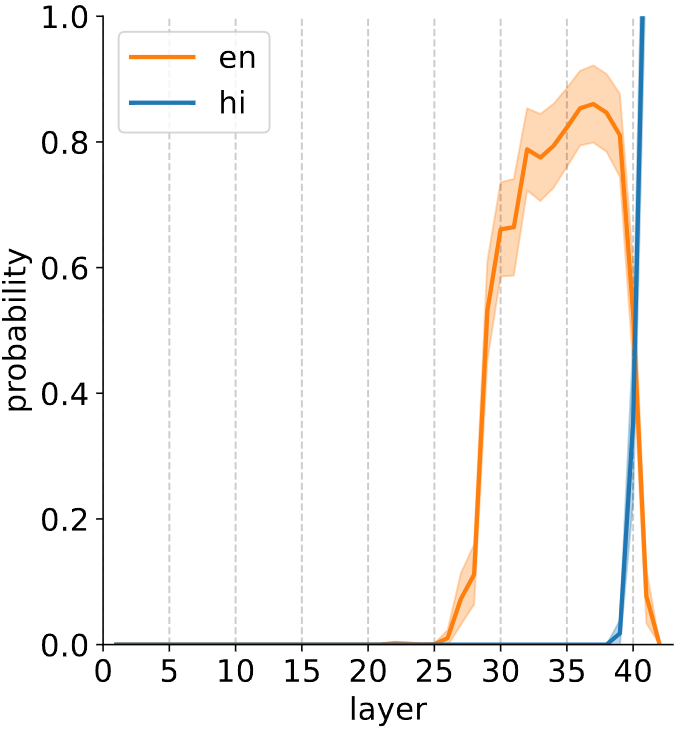}
        \caption{fr → hi}
        \label{fig:gemma 2 9b it fr-hi}
    \end{subfigure}

    \caption{\textbf{Language probabilities for latent layers} in translation from French to Telugu, Gujarati and Hindi in romanized (top row) and native scripts (bottom row) across various samples using the Gemma-2 9B IT model. On \textit{x}-axes, layer index; on \textit{y}-axes, probability (according to logit lens) of correct next token in a given language. Error bars represent 95\% Gaussian confidence intervals over input. In translations to non-English languages in romanized scripts (top row), target representations emerge slightly earlier—approximately one to two layers before layer 40—compared to their native script counterparts (bottom row), which only begin to appear from layer 40 onwards.}
    \label{fig:romanization early gemma 2 9b it fr appendix}
\end{figure*}

\begin{figure*}[htbp]
    \centering
    
    \begin{subfigure}{0.19\textwidth}
        \centering
        \includegraphics[width=\textwidth]{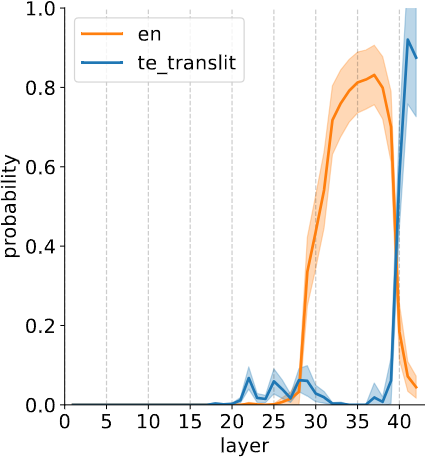}
        \caption{fr → te romanized}
        \label{fig:gemma 2 9b fr-te_translit}
    \end{subfigure}
    \hfill
    \begin{subfigure}{0.19\textwidth}
        \centering
        \includegraphics[width=\textwidth]{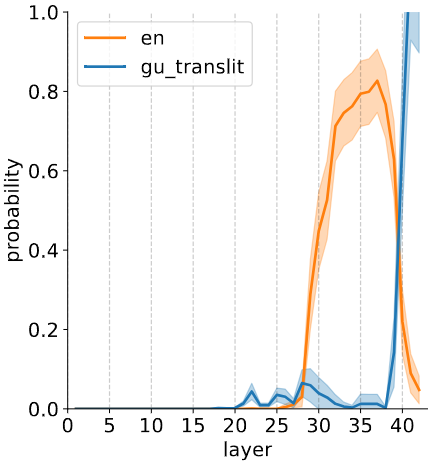}
        \caption{fr → gu romanized}
        \label{fig:gemma 2 9b fr-gu_translit}
    \end{subfigure}
    \hfill
    \begin{subfigure}{0.19\textwidth}
        \centering
        \includegraphics[width=\textwidth]{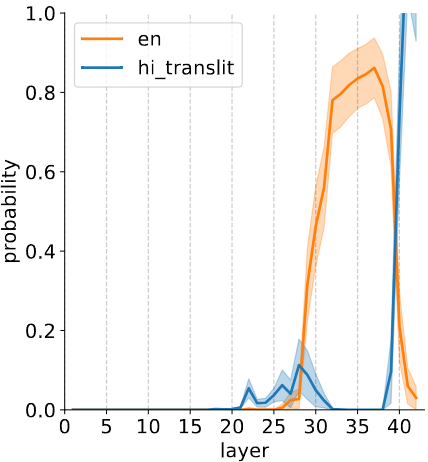}
        \caption{fr → hi romanized}
        \label{fig:gemma 2 9b fr-hi_translit}
    \end{subfigure}
    \hfill
    \begin{subfigure}{0.19\textwidth}
        \centering
        \includegraphics[width=\textwidth]{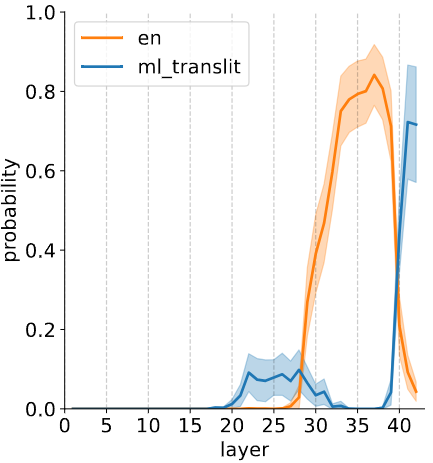}
        \caption{fr → ml romanized}
        \label{fig:gemma 2 9b fr-ml_translit}
    \end{subfigure}
    \hfill
    \begin{subfigure}{0.19\textwidth}
        \centering
        \includegraphics[width=\textwidth]{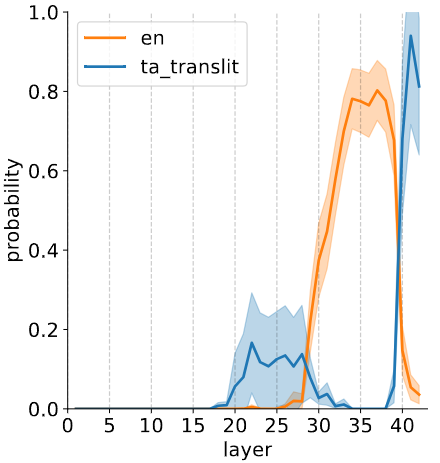}
        \caption{fr → ta romanized}
        \label{fig:gemma 2 9b fr-ta_translit}
    \end{subfigure}
    
    \vspace{1em}  
    
    \begin{subfigure}{0.19\textwidth}
        \centering
        \includegraphics[width=\textwidth]{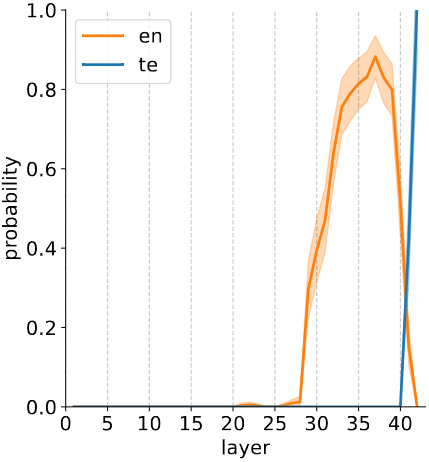}
        \caption{fr → te}
        \label{fig:gemma 2 9b fr-te}
    \end{subfigure}
    \hfill
    \begin{subfigure}{0.19\textwidth}
        \centering
        \includegraphics[width=\textwidth]{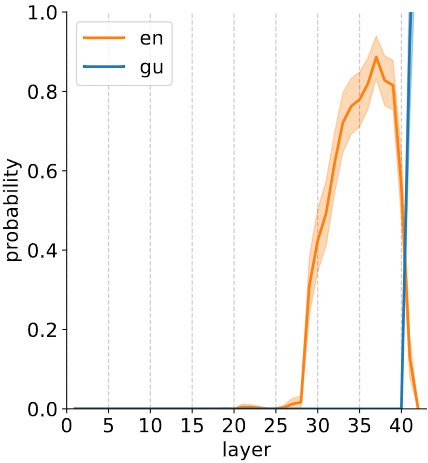}
        \caption{fr → gu}
        \label{fig:gemma 2 9b fr-gu}
    \end{subfigure}
    \hfill
    \begin{subfigure}{0.19\textwidth}
        \centering
        \includegraphics[width=\textwidth]{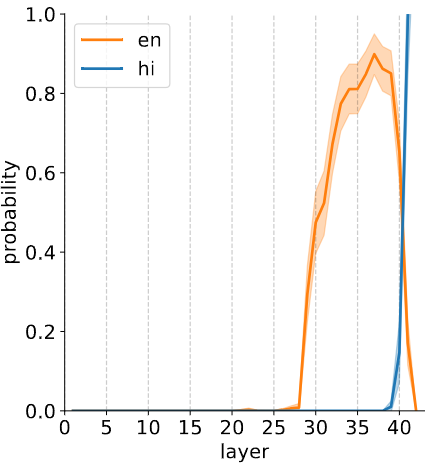}
        \caption{fr → hi}
        \label{fig:gemma 2 9b fr-hi}
    \end{subfigure}
    \hfill
    \begin{subfigure}{0.19\textwidth}
        \centering
        \includegraphics[width=\textwidth]{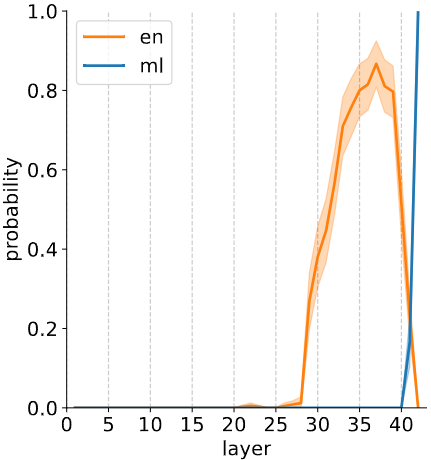}
        \caption{fr → ml}
        \label{fig:gemma 2 9b fr-ml}
    \end{subfigure}
    \hfill
    \begin{subfigure}{0.19\textwidth}
        \centering
        \includegraphics[width=\textwidth]{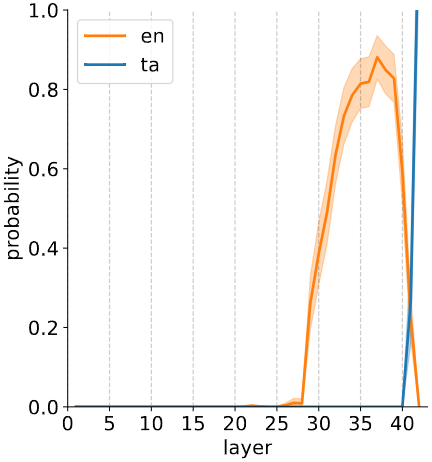}
        \caption{fr → ta}
        \label{fig:gemma 2 9b fr-ta}
    \end{subfigure}

    \caption{\textbf{Language probabilities for latent layers} in translation from French to five Indic languages (Telugu, Gujarati, Hindi, Malayalam, and Tamil) in romanized (top row) and native scripts (bottom row) using the Gemma-2 9B model. On \textit{x}-axes, layer index; on \textit{y}-axes, probability of correct next token in a given language. Error bars represent 95\% Gaussian confidence intervals over input. In translations using romanized scripts (top row), target representations emerge approximately 1-2 layers earlier than their native script counterparts (bottom row).}
    \label{fig:romanization early gemma 2 9b fr appendix}
\end{figure*}

\begin{figure*}[htbp]
    \centering
    
    \begin{subfigure}{0.19\textwidth}
        \centering
        \includegraphics[width=\textwidth]{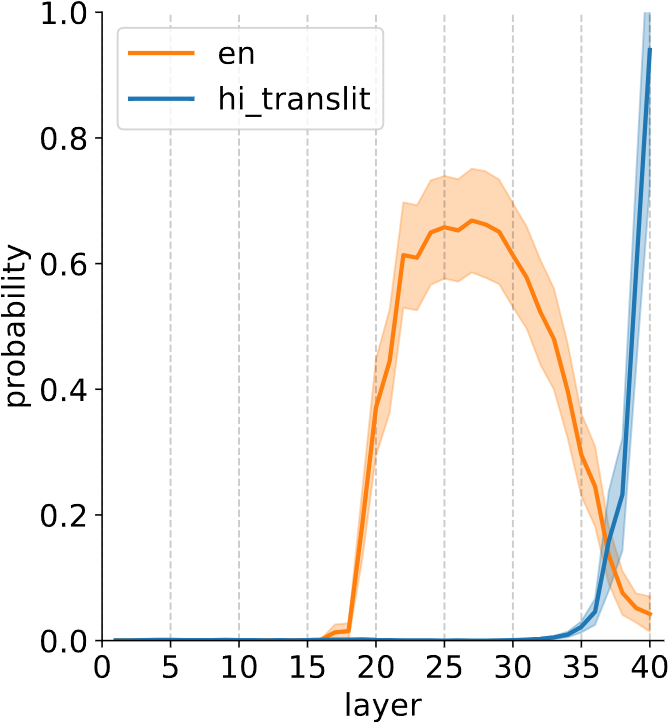}
        \caption{fr → hi romanized}
        \label{fig:llama 2 13b fr-hi_translit}
    \end{subfigure}
    \hfill
    \begin{subfigure}{0.19\textwidth}
        \centering
        \includegraphics[width=\textwidth]{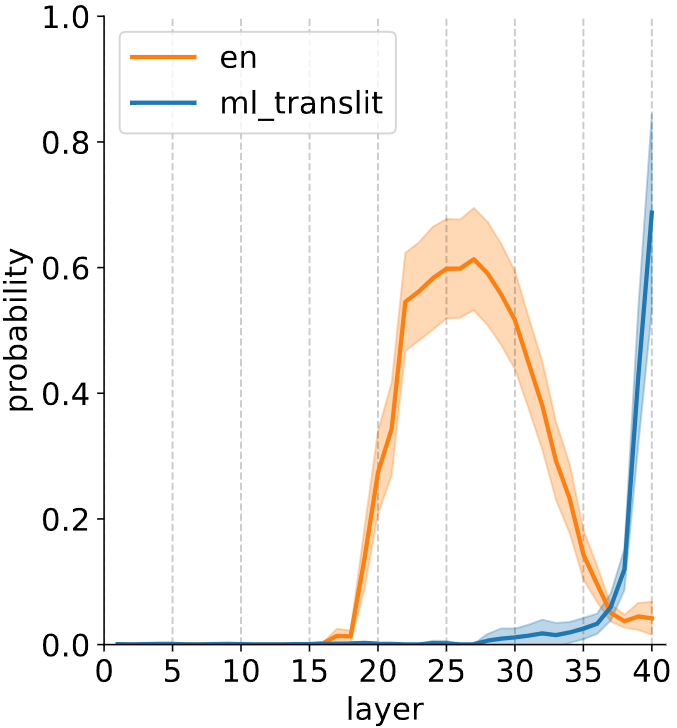}
        \caption{fr → ml romanized}
        \label{fig:llama 2 13b fr-ml_translit}
    \end{subfigure}
    \hfill
    \begin{subfigure}{0.19\textwidth}
        \centering
        \includegraphics[width=\textwidth]{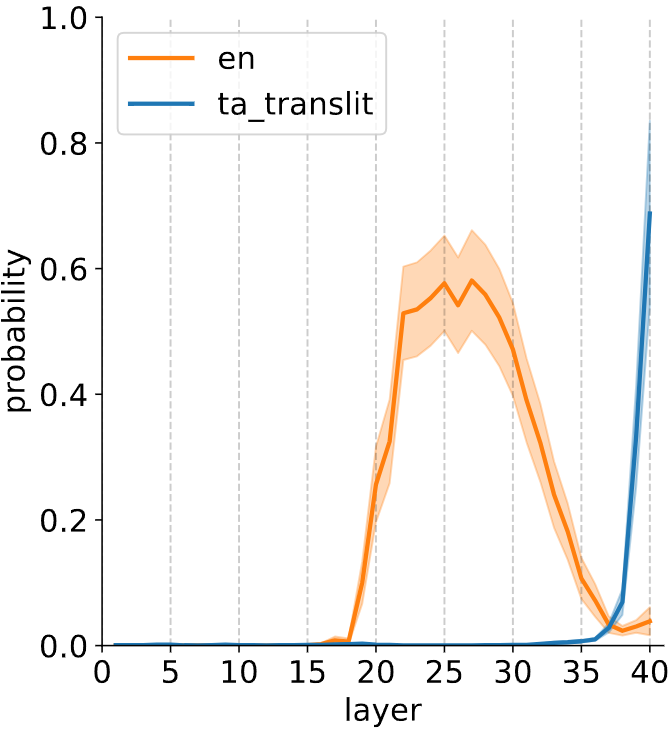}
        \caption{fr → ta romanized}
        \label{fig:llama 2 13b fr-ta_translit}
    \end{subfigure}
    
    \vspace{1em}  
    
    \begin{subfigure}{0.19\textwidth}
        \centering
        \includegraphics[width=\textwidth]{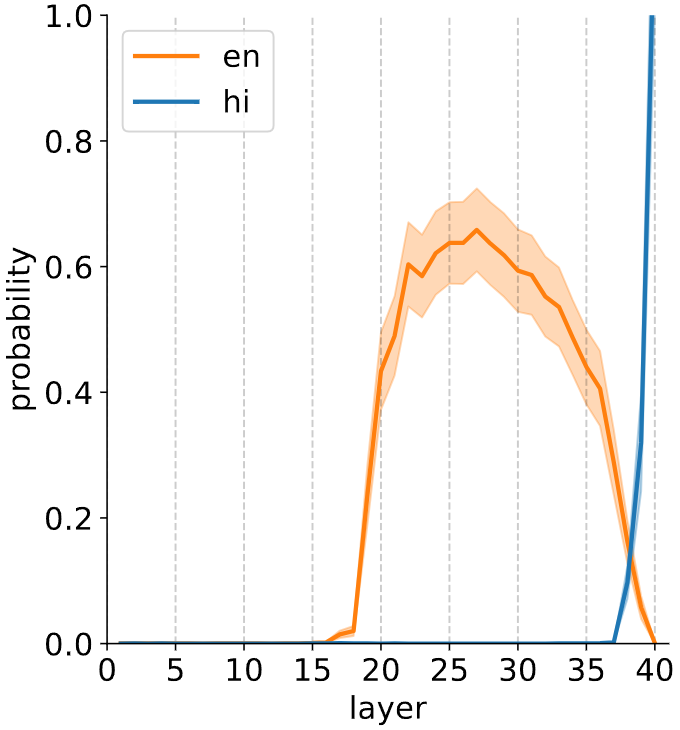}
        \caption{fr → hi}
        \label{fig:llama 2 13b fr-hi}
    \end{subfigure}
    \hfill
    \begin{subfigure}{0.19\textwidth}
        \centering
        \includegraphics[width=\textwidth]{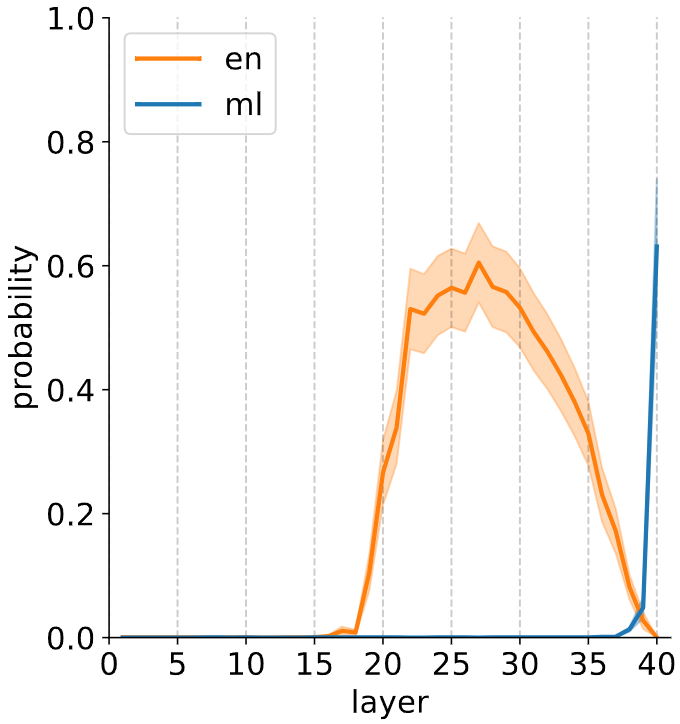}
        \caption{fr → ml}
        \label{fig:llama 2 13b fr-ml}
    \end{subfigure}
    \hfill
    \begin{subfigure}{0.19\textwidth}
        \centering
        \includegraphics[width=\textwidth]{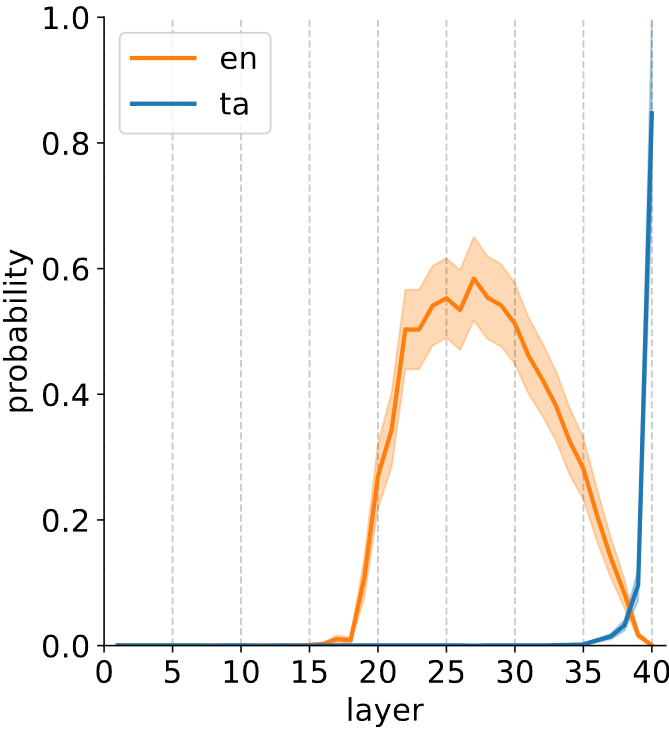}
        \caption{fr → ta}
        \label{fig:llama 2 13b fr-ta}
    \end{subfigure}

    \caption{\textbf{Language probabilities for latent layers} in translation from French to  Indic languages (Hindi, Malayalam, and Tamil) in romanized (top row) and native scripts (bottom row) using the Llama-2 13b model. On \textit{x}-axes, layer index; on \textit{y}-axes, probability of correct next token in a given language. Error bars represent 95\% Gaussian confidence intervals over input. In translations using romanized scripts (top row), target representations emerge approximately 1-2 layers earlier than their native script counterparts (bottom row).}
    \label{fig:romanization early llama 2 13b fr appendix}
\end{figure*}

Translation towards native script is compared with translation towards romanized script for gemma 2 9b it, gemma 2 9b and llama 2 13b models (see Figures \ref{fig:romanization early gemma 2 9b it fr appendix} to \ref{fig:romanization early llama 2 13b fr appendix}).

\section{Other Models: Mistral}
\label{sec:mistral}
\begin{figure*}[htbp]
    \centering
    \begin{tabular}{@{}c@{\hspace{1em}}c@{\hspace{1em}}c@{}}
        \multicolumn{1}{c}{All tokens} &
        \multicolumn{1}{c}{First token} &
        \multicolumn{1}{c}{Last token} \\[1ex]

        \includegraphics[width=0.3\textwidth]{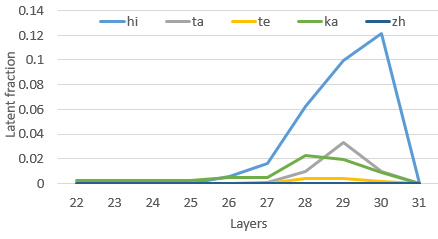} &
        \includegraphics[width=0.3\textwidth]{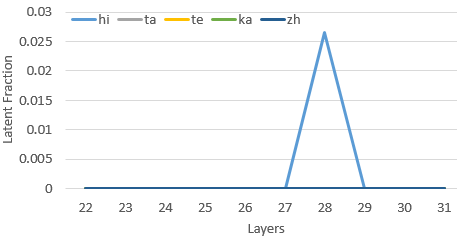} &
        \includegraphics[width=0.3\textwidth]{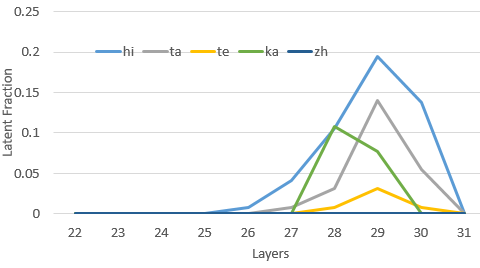} \\
    \end{tabular}
    
    Translation
    
    \vspace{2ex} 
    
    \begin{tabular}{ccc}
        \includegraphics[width=0.3\textwidth]{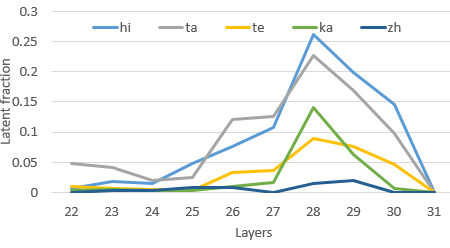} &
        \includegraphics[width=0.3\textwidth]{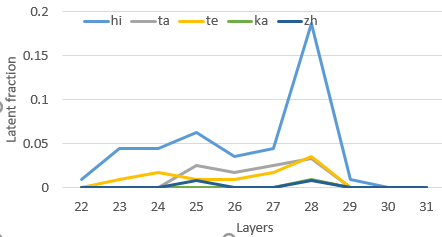} &
        \includegraphics[width=0.3\textwidth]{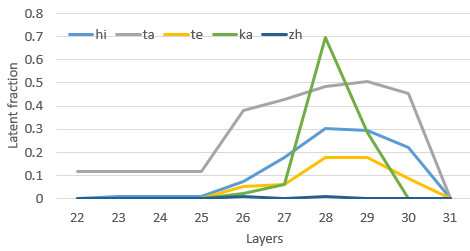} \\
    \end{tabular}
    
    Repetition

    \vspace{2ex} 
    
    \begin{tabular}{ccc}
        \includegraphics[width=0.3\textwidth]{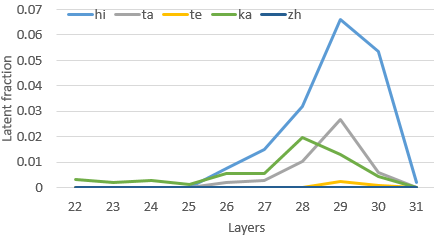} &
        \includegraphics[width=0.3\textwidth]{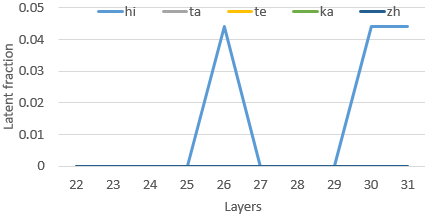} &
        \includegraphics[width=0.3\textwidth]{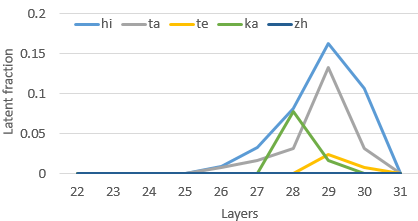} \\
    \end{tabular}
    
    Cloze
    
    \caption{\textbf{Distribution of Romanized Tokens Across Model Layers: Analysis of First, Last, and All Generation Timesteps.} This distribution is plotted across the last 10 layers of \textbf{Mistral-7B} model for initial, final and all token generation steps (columns)  for (a) translation task from English, (b) Repetition task, (c) Cloze task (rows) and is averaged across 100+ samples. \textit{X}-axes represents layer index, \textit{y}-axes represents latent fraction i.e. the fraction of timesteps where romanized tokens occur with a probability > 0.1 averaged over samples for a specific layer. We plot the distributions for  Tamil (ta), Telugu (te), Hindi (hi), Georgian (ka) and Chinese (zh).}
    \label{fig:mistral_latent_fraction_appendix}
\end{figure*}

We also perform our experiments on Mistral-7B \cite{jiang2023mistral}, a popular LLM known for its performance and efficiency. Layerrwise distribution of romanized tokens for initial, final and all token generation steps are presented in Figure \ref{fig:mistral_latent_fraction_appendix}.

\end{document}